\pdfoutput=1

\documentclass[11pt]{article}

\newcommand{\myheart}{\textsuperscript{$\heartsuit$}}
\newcommand{\myspadesuit}{\textsuperscript{$\spadesuit$}}

\usepackage[final]{acl}

\usepackage{times}
\usepackage{latexsym}

\usepackage[T1]{fontenc}

\usepackage[utf8]{inputenc}

\usepackage{microtype}
\usepackage{booktabs}
\usepackage{inconsolata}
\usepackage{graphicx}
\usepackage{multirow}
\usepackage{CJKutf8} 
\usepackage{makecell}
\usepackage{stfloats}
\usepackage{placeins}
\usepackage[many]{tcolorbox}
\tcbuselibrary{listings,breakable}
\newtcblisting{promptbox}[1][]{
  listing only,
  breakable,
  enhanced,
  colback=gray!4,
  colbacktitle=gray!18,
  colframe=black!45,
  coltitle=black,
  fonttitle=\bfseries\scriptsize,
  boxrule=0.35pt,
  arc=1pt,
  left=3pt,
  right=3pt,
  top=3pt,
  bottom=3pt,
  listing options={
    basicstyle=\fontsize{8.0}{8.35}\selectfont\ttfamily,
    breaklines=true,
    breakatwhitespace=true,
    columns=fullflexible,
    keepspaces=true,
    showstringspaces=false
  },
  #1
}
\tcbset{
  closehigh/.style={
    boxrule=0pt,
    colback=green!30,
    arc=3pt,
    boxsep=0pt,
    left=2.5pt,
    right=2.5pt,
    top=2.5pt,
    bottom=2.5pt,
    enhanced,
  }
}
\newcommand{\closehigh}[1]{\tcbox[closehigh, on line]{\textbf{#1}}}

\tcbset{
  ambihigh/.style={
    boxrule=0pt,
    colback=cyan!30,
    arc=3pt,
    boxsep=0pt,
    left=2.5pt,
    right=2.5pt,
    top=2.5pt,
    bottom=2.5pt,
    enhanced,
  }
}
\newcommand{\ambihigh}[1]{\tcbox[ambihigh, on line]{\textbf{#1}}}

\title{VIDA: A Dataset for Visually Dependent Ambiguity in Multimodal Machine Translation}

\author{\myspadesuit Jingheng Pan \and \myspadesuit \myheart Xintong Wang\thanks{Corresponding authors.} \\
 \myheart \textbf{Longyue Wang} \and \myheart \textbf{Liang Ding} \and \myheart \textbf{Weihua Luo} \and \myspadesuit \textbf{Chris Biemann}\footnotemark[1] \\
        \myspadesuit Department of Informatics, Universität Hamburg \quad \myheart Alibaba Group \\
        {\tt\small \myspadesuit\{jingheng.pan, xintong.wang, chris.biemann\}@uni-hamburg.de} \\
        {\tt\small \myheart liangding.liam@gmail.com},{\tt\small \myheart \{wanglongyue.wly, weihua.luowh\}@alibaba-inc.com} \\
        \footnotesize{\textit{Dataset:} \textcolor{blue}{\url{https://huggingface.co/datasets/p1k0/visually-dependent-ambiguity}}}
        }

\begin{document}
\maketitle
\begin{abstract}


Ambiguity resolution is a key challenge in multimodal machine translation (MMT), where models must genuinely leverage visual input to map an ambiguous expression to its intended meaning. 
Although prior work has proposed disambiguation-oriented benchmarks probing the role of vision, 
we observe that existing benchmarks remain limited by task-format mismatch, narrow ambiguity coverage, or insufficient visual-dependency validation.
Moreover, existing ambiguity evaluations are not well suited to diverse ambiguity types in open-ended translation. To address these limitations,
we present \textbf{VIDA} (Visually-Dependent Ambiguity), a dataset of 2,500 carefully curated instances in which resolving an annotated source span requires visual evidence. 
We further propose \textbf{Disambiguation-Centric Metrics} that use an LLM-as-a-judge classifier to verify whether annotated ambiguous expressions are resolved correctly at the span level. 
Evaluations with stronger recent LVLMs show that visual disambiguation remains challenging. Using chain-of-thought supervised fine-tuning as a diagnostic setting, 
we observe stronger out-of-distribution disambiguation than with SFT, with robust gains on collective-noun ambiguities and model-dependent gains on sentence-level ambiguities.

\end{abstract}

\section{Introduction}


Multimodal machine translation (MMT) extends neural machine translation by incorporating visual context to improve translation quality \citep{lala2018multimodal,yao-wan-2020-multimodal}. Recent Large Vision Language Models (LVLMs) show impressive performance on MMT benchmarks \citep{bai2025qwen2,zhu2025internvl3}. 
However, it remains unclear whether LVLMs truly leverage visual information during MMT. Earlier studies \citep{elliott-2018-adversarial, wu-etal-2021-good} showed that replacing or perturbing images often leads to only minor degradations, raising questions about the actual contribution of the visual modality. This concern remains relevant in the LVLM era: stronger visual perception does not necessarily guarantee reliable use of visual evidence during translation.



Ambiguity resolution provides a direct probe of visual dependence in MMT, and existing work has made progress in this direction by introducing multimodal disambiguation benchmarks \citep{ma-etal-2024-3am,mma,futeral-etal-2023-tackling}. 
Yet no existing benchmark combines open-ended MMT generation with broad ambiguity coverage and visual-dependency validation.
CoMMuTE \citep{futeral-etal-2023-tackling} is a contrastive benchmark for word-level and gender ambiguity that scores candidate translations rather than requiring models to generate translations freely. 
3AM \citep{ma-etal-2024-3am} targets English--Chinese MMT and primarily focuses on word-level ambiguity, but includes many instances resolvable from text alone. MMA \citep{mma} covers sentence-level ambiguity in a VQA-style format rather than translation generation. Additionally, prior work on MMT ambiguity evaluation spans several paradigms, including contrastive evaluation with predefined translation candidates \citep{futeral-etal-2023-tackling} and rule-based matching with lexical variants \citep{lala2018multimodal, li-etal-2021-vision}. While these settings provide ambiguity-oriented evaluation, they are less suitable for open-ended MMT generation with broader ambiguity types and valid paraphrases. General MT metrics such as BLEU and COMET assess overall translation quality, but do not directly indicate whether a particular ambiguous span has been resolved correctly.

Reliable evaluation of visual disambiguation in MMT requires both an MMT dataset with visually-dependent ambiguities and an evaluation method that directly measures disambiguation accuracy.
We therefore introduce \textbf{VIDA} (\textbf{Vi}sually-\textbf{D}ependent \textbf{A}mbiguity), an MMT dataset of 2,500 instances capturing visually-dependent translation ambiguities at the word and sentence levels, as well as collective-noun phenomena. We further propose \textbf{Disambiguation-Centric Metrics} based on an LLM-as-a-judge classifier. These metrics complement standard translation metrics by explicitly targeting span-level disambiguation accuracy. 
Even recent LVLMs such as Qwen3.5-27B~\citep{qwen35} resolve all annotated ambiguities in fewer than half of the test instances.


Beyond dataset and evaluation, we use chain-of-thought supervised fine-tuning (CoT-SFT) \citep{muennighoff2025s1} with a manually designed template as a diagnostic setting to test whether explicit visual reasoning supervision improves disambiguation beyond standard SFT. Experiments show robust gains on VIDA-CollN across models, while the effect on VIDA-Sent is model-dependent. The main contributions of this paper are:


\begin{itemize}

\item We introduce \textbf{VIDA}, an MMT dataset featuring visually-dependent ambiguities at both word and sentence levels. 

\item We propose \textbf{Disambiguation-Centric Metrics} to directly measure disambiguation accuracy, complementing standard translation metrics for MMT disambiguation.

\item We use CoT-SFT as a diagnostic setting to examine whether explicit visual-reasoning supervision improves disambiguation beyond standard SFT.
\end{itemize}

\section{Related Work}

\paragraph{Multimodal Disambiguation Datasets.}
Prior work has studied visual disambiguation through translation-based benchmarks \citep{lala2018multimodal,li-etal-2021-vision,ma-etal-2024-3am}, contrastive MMT benchmarks \citep{futeral-etal-2023-tackling}, and VQA-style ambiguity benchmarks \citep{mma}. Building on such benchmarks, DeDiT \citep{liu-etal-2025-detect} examines explicit visual reasoning in MMT and evaluates cross-dataset transfer from Multi30K to CoMMuTE. In comparison, VIDA targets open-ended MMT generation with broader ambiguity coverage, retaining only cases whose correct interpretation requires visual evidence. Sharing the motivation of DeDiT, we use CoT-SFT as a diagnostic setting, but since VIDA separates ambiguity types into subsets, we test whether visual reasoning transfers from word-level ambiguities to held-out sentence-level and collective-noun ambiguities rather than across datasets.

\paragraph{Disambiguation Metrics for MMT.}
Prior work has evaluated multimodal ambiguity with contrastive candidate ranking \citep{futeral-etal-2023-tackling}, rule-based matching over predefined lexical variants \citep{lala2018multimodal, li-etal-2021-vision}, and general MT metrics \citep{papineni-etal-2002-bleu,rei-etal-2020-comet}. These approaches are less suitable for open-ended MMT generation with broader ambiguity types, where correct disambiguation may be expressed through paraphrases or lexical variation. We therefore propose Disambiguation-Centric Metrics using an LLM-as-a-judge classifier to directly assess span-level disambiguation accuracy. A detailed comparison is provided in Appendix~\ref{sec:positioning}.

\begin{figure}[t]
    \centering
  \includegraphics[width=0.95\linewidth]{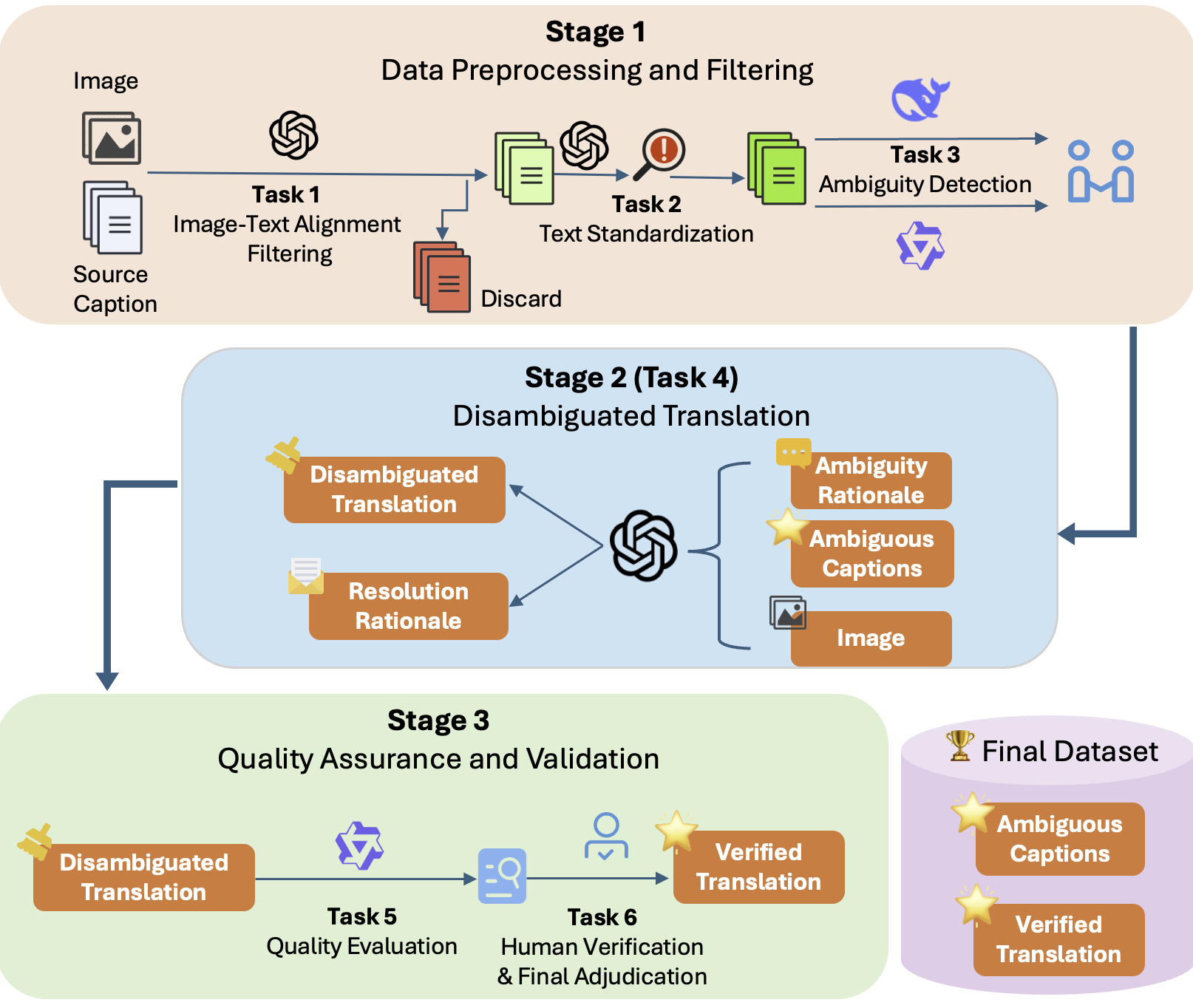}
  \caption{Three-stage VIDA curation pipeline}\label{fig:pipeline}
\end{figure}

\section{Dataset Curation} \label{sec:pipeline}


To construct an MMT dataset with visually-dependent translation ambiguities, we adopt a three-stage semi-automatic pipeline, as shown in \autoref{fig:pipeline}. 
Our goal is to collect instances where the disambiguating evidence is grounded in the image, and to provide human-verified reference translations.
VIDA is built on two existing resources: VIDA-Base and VIDA-CollN are curated from 3AM \citep{ma-etal-2024-3am}, and VIDA-Sent is adapted from MMA \citep{mma}.


\subsection{Stage 1: Data Preprocessing and Filtering}

The goal of this stage is to curate image--text aligned and textually clean source captions with visually-dependent ambiguities. 
We use GPT-4o \citep{openai2025gpt4o} to filter mismatched image--text pairs and normalize text. 
We then apply a dual-model consensus strategy with two state-of-the-art commercial LLMs, Qwen-Max \citep{qwen25} and DeepSeek-v3 \citep{liu2024deepseek}: each model independently assesses whether the caption remains ambiguous without the image. 
We retain only captions that both models judge as ambiguous, and record an \textit{Ambiguous Caption} and an \textit{Ambiguity Rationale} that points to the ambiguous span and the disambiguation cue. 


\subsection{Stage 2: Disambiguated Translation} \label{sec:stage2}
In this stage, we produce high-quality disambiguated translations for each retained caption. 
For each caption from Stage 1, we generate a disambiguated translation using GPT-4o with a structured input that includes the \textit{Ambiguous Caption}, the paired \textit{Image}, and the \textit{Ambiguity Rationale} from the previous stage. The model outputs both a \textit{Disambiguated Translation} that resolves the ambiguity and a \textit{Resolution Rationale} explaining how visual information is used to support the decision.

\subsection{Stage 3: Quality Assurance and Validation}

This stage finalizes the dataset with an LLM-as-a-judge quality check followed by human verification, ensuring both translation quality and correct visual disambiguation.
We employ Qwen-Max to evaluate each disambiguated translation along two dimensions: semantic preservation and fluency. For each dimension, Qwen-Max outputs a score from 1 to 5 together with a brief justification. We flag cases with scores below 4 as potentially problematic and prioritize them for closer inspection.

To ensure translation correctness, we further conduct human verification with two native Chinese speakers (one Ph.D. and one M.S. in Computer Science). For each instance, annotators are shown the image, the source caption, the ambiguity rationale, the candidate translation, and Qwen-Max's judgment. 
Each annotator independently reviews the translation using three criteria: (i) whether the annotated ambiguity is correctly resolved, (ii) whether the translation is fluent, and (iii) whether the meaning of the translation is preserved. If the candidate fails any criterion, the annotator provides a corrected translation based on the image and the ambiguity rationale. During verification, we identified collective noun cases where underspecified terms must be concretized based on the image content and group these cases into the \textbf{Collective Noun Subset}. Finally, \textbf{VIDA} comprises 2{,}500 instances across three subsets (\textbf{VIDA-Base}, \textbf{VIDA-Sent}, and \textbf{VIDA-CollN}). See Appendix~\ref{sec:dataset_stat} for details.





\section{Evaluation Metrics for Disambiguation}



In MMT disambiguation, the objective of evaluation is to measure disambiguation accuracy rather than overall translation quality. 
We use an LLM-as-a-judge to classify span-level correctness and compute accuracy from binary outputs.
Specifically, we fine-tune Qwen3-8B \citep{yang2025qwen3} on \textbf{VIDA} to determine whether annotated ambiguous expressions are correctly resolved in translation. 
We train the classifier in a contrastive setting. Gold disambiguated translations are used as positive samples. Negative samples are candidate translations from our curation pipeline that fail to resolve the annotated ambiguity. 
Building on this, we introduce two complementary \textbf{Disambiguation-Centric Metrics} for a comprehensive evaluation of disambiguation performance:
\textbf{Disambi-Term} measures term-level disambiguation accuracy by evaluating each annotated ambiguous term in the entire dataset.
\textbf{Disambi-Inst.} reports instance-level accuracy, counting a sentence as correct only if all ambiguous expressions within it are correctly resolved. 
Judge reliability and training details are provided in Appendix~\ref{sec:appendix_validation}.

\begin{table*}[t]
\centering
\resizebox{\textwidth}{!}{%
\begin{tabular}{@{}lllccccccccc@{}}
\toprule
\textbf{Model} & \textbf{Dataset} & \textbf{Model Setting} & \textbf{BLEU} & \textbf{chrF} & \textbf{chrF++} & \textbf{TER} & \textbf{BERT-F1} & \textbf{METEOR} & \textbf{COMET} & \textbf{Disambi-Term} & \textbf{Disambi-Inst.} \\ \midrule

\multirow{12}{*}{\textbf{InternVL3-8B}} 
 & \multirow{3}{*}{All-Test} & Vanilla & 48.04 & 41.95 & 32.98 & \closehigh{40.29} & 86.63 & 58.47 & 84.49 & 50.86 & 39.81 \\
 & & SFT & \closehigh{49.20} & \closehigh{42.73} & 32.82 & 43.44 & 87.07 & \closehigh{58.95} & 85.55 & 54.36 & 43.77 \\
 & & CoT-SFT & 47.64 & 41.16 & \closehigh{32.99} & 41.61 & \closehigh{87.18} & 58.78 & \closehigh{85.88} & \ambihigh{58.45} & \ambihigh{48.78} \\ \cmidrule(l){2-12}

 & \multirow{3}{*}{VIDA-Base-Test} & Vanilla & 53.51 & 46.76 & 36.87 & 35.66 & 88.84 & 65.24 & 86.08 & 60.18 & 46.55 \\
 & & SFT & \closehigh{55.31} & \closehigh{48.31} & \closehigh{37.13} & \closehigh{34.28} & \closehigh{89.61} & \closehigh{66.55} & \closehigh{87.30} & 62.67 & 50.17 \\
 & & CoT-SFT & 51.10 & 44.41 & 35.75 & 38.06 & 88.56 & 63.25 & 86.44 & \ambihigh{64.89} & \ambihigh{51.38} \\ \cmidrule(l){2-12}

 & \multirow{3}{*}{VIDA-Sent} & Vanilla & 42.51 & 36.85 & 33.01 & \closehigh{44.76} & 84.31 & 52.54 & 84.21 & 50.00 & 50.00 \\
 & & SFT & 36.99 & 35.69 & 31.96 & 67.93 & 83.93 & 51.67 & 84.70 & 55.45 & 55.45 \\
 & & CoT-SFT & \closehigh{44.22} & \closehigh{38.19} & \closehigh{34.28} & 45.27 & \closehigh{85.70} & \closehigh{55.32} & \closehigh{86.39} & \ambihigh{58.97} & \ambihigh{58.97} \\ \cmidrule(l){2-12}

 & \multirow{3}{*}{VIDA-CollN} & Vanilla & 36.56 & 31.66 & 27.14 & 49.24 & 84.63 & 48.79 & 81.39 & 18.36 & 12.16 \\
 & & SFT & 37.97 & 32.92 & \closehigh{28.22} & 49.11 & 85.11 & 50.56 & 82.60 & 22.62 & 14.90 \\
 & & CoT-SFT & \closehigh{39.26} & \closehigh{33.89} & 25.70 & \closehigh{48.52} & \closehigh{85.70} & \closehigh{51.05} & \closehigh{83.96} & \ambihigh{38.36} & \ambihigh{32.55} \\ \midrule \midrule

\multirow{12}{*}{\textbf{Qwen2.5-VL-7B}} 
 & \multirow{3}{*}{All-Test} & Vanilla & 47.85 & 41.67 & 34.12 & 42.74 & 86.56 & 58.88 & 84.83 & 50.08 & 39.81 \\
 & & SFT & \closehigh{49.13} & \closehigh{42.85} & \closehigh{34.58} & \closehigh{40.34} & \closehigh{87.38} & \closehigh{59.51} & 85.82 & 52.81 & 42.46 \\
 & & CoT-SFT & 47.59 & 41.39 & 33.26 & 42.49 & 87.06 & 58.60 & \closehigh{85.84} & \ambihigh{55.51} & \ambihigh{46.08} \\ \cmidrule(l){2-12}

 & \multirow{3}{*}{VIDA-Base-Test} & Vanilla & 52.38 & 45.66 & 37.66 & 37.64 & 88.53 & 64.45 & 86.30 & 58.49 & 46.38 \\
 & & SFT & \closehigh{53.88} & \closehigh{47.09} & \closehigh{38.35} & \closehigh{36.17} & \closehigh{89.11} & \closehigh{65.48} & \closehigh{87.07} & \ambihigh{61.42} & \ambihigh{49.31} \\
 & & CoT-SFT & 50.41 & 44.57 & 36.00 & 39.44 & 88.32 & 62.75 & 86.35 & 60.71 & 46.90 \\ \cmidrule(l){2-12}

 & \multirow{3}{*}{VIDA-Sent} & Vanilla & 44.46 & 38.92 & 34.97 & 50.95 & 84.21 & 54.78 & 84.41 & 51.28 & 51.28 \\
 & & SFT & 45.12 & 39.52 & \closehigh{35.59} & \closehigh{42.66} & \closehigh{85.86} & \closehigh{55.41} & 86.06 & 52.56 & 52.56 \\
 & & CoT-SFT & \closehigh{45.54} & \closehigh{39.79} & 35.43 & 45.17 & 85.66 & 55.02 & \closehigh{86.41} & \ambihigh{60.26} & \ambihigh{60.26} \\ \cmidrule(l){2-12}

 & \multirow{3}{*}{VIDA-CollN} & Vanilla & 38.06 & 32.83 & 24.63 & 50.54 & 84.87 & 51.16 & 82.06 & 19.02 & 12.16 \\
 & & SFT & \closehigh{39.02} & \closehigh{33.71} & \closehigh{25.28} & \closehigh{49.24} & 85.30 & 50.96 & 82.69 & 21.31 & 14.51 \\
 & & CoT-SFT & 38.21 & 32.49 & 24.51 & 50.71 & \closehigh{85.32} & \closehigh{51.34} & \closehigh{83.39} & \ambihigh{33.77} & \ambihigh{27.45} \\ \bottomrule
\end{tabular}%
}
\caption{Performance comparison of \textbf{InternVL3-8B} and \textbf{Qwen2.5-VL-7B} under Vanilla, SFT, and CoT-SFT settings. \closehigh{Box} highlights best standard metrics, \ambihigh{Box} highlights best disambiguation metrics.}
\label{tab:combined_comparison}
\end{table*}

\section{Experiments} \label{sec:experiments}
\subsection{Experimental Settings}

\paragraph{Dataset}
We conduct experiments on the VIDA dataset. VIDA-Base is split 7:3 into 1,352 training and 580 test samples. VIDA-Sent and VIDA-CollN are held out from training as out-of-distribution (OOD) test sets due to limited size, which also allows evaluation on unseen ambiguity types.

\paragraph{Metrics}
We report standard MT metrics (\textit{BLEU}, \textit{chrF}/\textit{chrF++} \citep{popovic-2015-chrf,popovic-2017-chrf}, \textit{METEOR} \citep{banerjee-lavie-2005-meteor}, \textit{TER} \citep{snover-etal-2006-study}, \textit{BERT-F1} \citep{devlin2019bert}, and \textit{COMET}) to assess overall translation quality, together with Disambi-Term and Disambi-Inst.\ for disambiguation accuracy. To assess the reliability of gains on the small OOD subsets, we estimate 95\% CIs using paired bootstrap for Disambi-Inst.\ and instance-level cluster bootstrap for Disambi-Term, and 
test instance-level significance with exact McNemar tests (Appendix~\ref{sec:stat_details}).

\paragraph{Models and Baselines}
We evaluate \textbf{Qwen2.5-VL-7B} \citep{bai2025qwen2} and \textbf{InternVL3-8B} \citep{zhu2025internvl3} under three settings: vanilla inference, supervised fine-tuning (SFT), and chain-of-thought supervised fine-tuning (CoT-SFT) with manually synthesized ambiguity-resolution traces (Appendix~\ref{sec:cot-sft}). We additionally evaluate Qwen3.5-27B and Gemma-4-26B-A4B \citep{gemma4} under vanilla inference and extend the SFT/CoT-SFT comparison to InternVL3-14B to examine CoT-SFT's effectiveness at a larger model scale.

\subsection{Experimental Results}


\paragraph{Analysis on In-Domain Dataset.} Evaluation on the in-domain test set (\textbf{VIDA-Base-Test}) in \autoref{tab:combined_comparison} examines how well the models fit the training distribution (VIDA-Base-Train).
The SFT setting achieves the strongest overall translation quality for both models across most standard translation metrics. 
CoT-SFT remains competitive on semantic metrics (e.g., COMET) for both models, but can slightly underperform SFT on surface-overlap metrics such as BLEU, likely because ambiguity resolution often requires paraphrasing or structural reformulation that deviates from reference wording. To further assess whether these surface-metric gaps indicate an overall quality loss, we use an LVLM judge (Qwen3.6-Plus; \citealp{qwen36plus}) and find CoT-SFT comparable to SFT (Appendix~\ref{sec:llm_quality}). Under Disambiguation-Centric evaluation, CoT-SFT yields the best performance for InternVL3-8B (Disambi-Term/Inst.\ \textbf{64.89/51.38}), surpassing both SFT and Vanilla. In contrast, for Qwen2.5-VL-7B, CoT-SFT improves over Vanilla but remains slightly below SFT (60.71/46.90 vs.\ 61.42/49.31), suggesting that in-domain gains from reasoning supervision are model-dependent. We attribute part of the CoT-SFT disambiguation gap to overthinking behaviors and provide quantitative and qualitative analyses in Appendix~\ref{sec:error_analysis}.

\paragraph{Analysis on Out-of-Distribution Dataset.} Evaluation on the out-of-distribution (OOD) datasets (\textbf{VIDA-Sent} and \textbf{VIDA-CollN}) examines whether models can generalize to unseen ambiguity types. 
Compared to in-domain results, CoT-SFT shows clearer advantages under distribution shift. On standard translation metrics, CoT-SFT achieves the highest COMET on both OOD subsets for both models, indicating stronger semantic adequacy. 
Disambiguation-centric metrics make the advantage more explicit. On VIDA-Sent, CoT-SFT achieves consistent gains over SFT (approximately \textbf{+3.5} points for InternVL3-8B and \textbf{+7.7} points for Qwen2.5-VL-7B on both Disambi-Term and Disambi-Inst.). The gains are substantially larger on VIDA-CollN (over \textbf{+15} points for InternVL3-8B and \textbf{+12} points for Qwen2.5-VL-7B), suggesting that CoT-SFT demonstrates stronger generalization beyond the training distribution.


\paragraph{Statistical Significance of OOD Gains.}
Because the OOD subsets are small, we further test whether these gains are statistically reliable (\autoref{tab:ood_significance}). Three of the four instance-level gains are significant at $\alpha=0.05$ and remain significant after Bonferroni correction ($\alpha=0.0125$): Qwen2.5-VL-7B on VIDA-Sent and both models on VIDA-CollN, and the term-level CIs lead to the same conclusion. The gain for InternVL3-8B on VIDA-Sent is positive but not significant ($p=0.1352$). These results indicate that the benefit of CoT-SFT on VIDA-CollN is significant for both models and consistent at the term and instance levels. In contrast, the benefit on VIDA-Sent is model-dependent, possibly because sentence-level ambiguities such as idioms and pragmatic interpretations differ more from the word-level training instances, so the learned reasoning patterns transfer less uniformly across models.

\begin{table}[t]
\centering
\resizebox{\columnwidth}{!}{%
\begin{tabular}{lccc}
\toprule
\textbf{Subset} & \textbf{$\Delta$ Inst. [95\% CI]} & \textbf{McNemar $p$} & \textbf{$\Delta$ Term [95\% CI]} \\
\midrule
\multicolumn{4}{@{}l}{\textit{Qwen2.5-VL-7B}} \\
VIDA-Sent  & \textbf{+7.69 [3.21, 12.18]} & 0.0012 & \textbf{+7.69 [3.21, 12.18]} \\
VIDA-CollN & \textbf{+12.94 [7.84, 18.04]} & $<$0.00001 & \textbf{+12.46 [7.33, 17.78]} \\
\midrule
\multicolumn{4}{@{}l}{\textit{InternVL3-8B}} \\
VIDA-Sent  & +3.53 [$-$0.64, 8.01] & 0.1352 & +3.53 [$-$0.64, 8.01] \\
VIDA-CollN & \textbf{+17.65 [12.16, 23.14]} & $<$0.00001 & \textbf{+15.74 [10.36, 21.38]} \\
\bottomrule
\end{tabular}%
}
\caption{CoT-SFT vs.\ SFT on the OOD subsets. $\Delta$: CoT-SFT minus SFT; $p$: exact McNemar test. Bold: 95\% CI excludes zero. }
\label{tab:ood_significance}
\end{table}

\paragraph{Analysis on All-Test Dataset.} All-Test pools all subsets into a single evaluation rather than averaging per-subset scores, yielding an overall score weighted by subset size. On standard translation metrics, SFT generally improves over Vanilla for both models, indicating better overall translation quality on the mixed test distribution. In contrast, CoT-SFT consistently yields higher semantic adequacy and disambiguation accuracy than SFT: it achieves the highest COMET and improves Disambi-Term/Inst.\ by about  \textbf{+4.1/+5.0} for InternVL3-8B and \textbf{+2.7/+3.6} for Qwen2.5-VL-7B. Overall, CoT-SFT trades a small loss in surface overlap for higher semantic adequacy and disambiguation accuracy, with most of the disambiguation gain coming from the OOD subsets.

\subsection{Stronger LVLMs and Larger-Scale CoT-SFT}
\label{sec:stronger_models}

\begin{table}[t]
\centering
\resizebox{\columnwidth}{!}{%
\begin{tabular}{@{}llcccc@{}}
\toprule
\textbf{Model} & \textbf{Setting} & \textbf{BLEU} & \textbf{COMET} & \makecell{\textbf{Disambi-}\\\textbf{Term}} & \makecell{\textbf{Disambi-}\\\textbf{Inst.}} \\ \midrule
\textbf{Qwen3.5-27B} & Vanilla & 48.14 & 86.38 & 57.25 & 46.47 \\
\textbf{Gemma-4-26B-A4B} & Vanilla & 45.82 & 85.53 & 54.86 & 44.46 \\ \midrule
\multirow{3}{*}{\textbf{InternVL3-14B}}
 & Vanilla & 49.93 & 84.90 & 50.28 & 39.67 \\
 & SFT & \closehigh{51.23} & 85.75 & 52.97 & 42.63 \\
 & CoT-SFT & 49.55 & \closehigh{86.20} & \ambihigh{58.47} & \ambihigh{50.04} \\ \bottomrule
\end{tabular}%
}
\caption{All-Test results of stronger LVLMs (top) and InternVL3-14B (bottom). Highlighting follows \autoref{tab:combined_comparison}; full results are in Appendix~\ref{sec:full_results}.}
\label{tab:stronger_models}
\end{table}

\paragraph{Evaluation with Stronger LVLMs.}
As shown in \autoref{tab:stronger_models}, Qwen3.5-27B and Gemma-4-26B-A4B achieve higher disambiguation accuracy than the vanilla 7B/8B models in \autoref{tab:combined_comparison} (up to 57.25/46.47 vs.\ 50.86/39.81 Disambi-Term/Inst.). Nevertheless, fewer than half of the All-Test instances have all annotated ambiguities correctly resolved, indicating that visual disambiguation on VIDA is not solved by stronger recent LVLMs. Notably, InternVL3-8B with CoT-SFT (48.78 Disambi-Inst.) already exceeds both, although it is trained on VIDA-Base. The subset-level results (Appendix~\ref{sec:full_results}) show that the remaining difficulty is most pronounced on VIDA-CollN, where both models reach at most 16.02 Disambi-Inst., close to the vanilla 7B/8B models (12.16) and far below CoT-SFT (27.45--32.55).

\paragraph{Larger-Scale CoT-SFT Evaluation.}
On InternVL3-14B, CoT-SFT improves Disambi-Term/Inst.\ by 5.50/7.41 points over SFT and achieves the highest COMET, while SFT retains the highest BLEU. This reproduces the pattern observed at the 7B/8B scale: SFT better fits the surface form of the references, whereas CoT-SFT improves semantic adequacy and disambiguation. As at the smaller scale, the gains come mainly from the OOD subsets (+9.30 and +20.00 Disambi-Inst.\ on VIDA-Sent and VIDA-CollN; Appendix~\ref{sec:full_results}), indicating that the benefit of CoT-SFT extends beyond the 7B/8B scale.



\section{Conclusion}
We introduce VIDA, an MMT dataset for visually-dependent word- and sentence-level ambiguities, and propose Disambiguation-Centric Metrics that directly measure disambiguation accuracy based on an LLM-as-a-judge classifier. 
Evaluations with stronger recent LVLMs show that VIDA remains challenging, especially for collective nouns. Our comparison of SFT and CoT-SFT shows that SFT better fits the surface patterns of the training references, whereas CoT-SFT improves visually grounded disambiguation. The gains are robust on VIDA-CollN across both models, while the effect on VIDA-Sent is model-dependent;
the disambiguation benefit also holds for InternVL3-14B.


\clearpage

\section*{Limitations}
Although \textbf{VIDA} targets highly visually dependent ambiguities, the dataset remains modest in scale compared to large-scale translation corpora and is currently limited to English--Chinese translation. In addition, the reasoning traces used to supervise CoT-SFT are synthetically constructed based on manually designed patterns, which may not reflect natural model reasoning. Although we include stronger vanilla baselines and a 14B experiment, the CoT-SFT evaluation still covers a limited number of open-weight model families.
Future work includes expanding the dataset to additional language pairs, validating the CoT-SFT findings on a broader range of LVLMs, and moving beyond fixed templates toward dynamic or adaptive reasoning generation, for instance through reinforcement learning or preference-based optimization methods that let the model adapt its reasoning depth to the difficulty of the ambiguity.

\section*{Ethics Statement}

The proposed \textbf{VIDA} dataset is curated from publicly available 
data sources and inherits their research-use license; 
the images do not introduce additional personally identifying 
information beyond the underlying datasets. Annotators were informed of the task goals before annotation. 
The dataset is publicly released with usage guidelines to support research on multimodal machine translation and ambiguity resolution.


\bibliography{custom_mandatory}

@inproceedings{lala2018multimodal,
  title={Multimodal lexical translation},
  author={Lala, Chiraag and Specia, Lucia},
  booktitle={Proceedings of the Eleventh International Conference on Language Resources and Evaluation (LREC 2018)},
  year={2018},
  url={https://aclanthology.org/L18-1602.pdf}
}

@inproceedings{guo-etal-2022-lvp,
    title = "{LVP}-{M}3: Language-aware Visual Prompt for Multilingual Multimodal Machine Translation",
    author = "Guo, Hongcheng  and
      Liu, Jiaheng  and
      Huang, Haoyang  and
      Yang, Jian  and
      Li, Zhoujun  and
      Zhang, Dongdong  and
      Cui, Zheng",
    booktitle = "Proceedings of the 2022 Conference on Empirical Methods in Natural Language Processing",
    month = dec,
    year = "2022",
    address = "Abu Dhabi, United Arab Emirates",
    publisher = "Association for Computational Linguistics",
    url = "https://aclanthology.org/2022.emnlp-main.184/",
    doi = "10.18653/v1/2022.emnlp-main.184",
    pages = "2862--2872"
}

@inproceedings{yao-wan-2020-multimodal,
    title = "Multimodal Transformer for Multimodal Machine Translation",
    author = "Yao, Shaowei  and
      Wan, Xiaojun",
    editor = "Jurafsky, Dan  and
      Chai, Joyce  and
      Schluter, Natalie  and
      Tetreault, Joel",
    booktitle = "Proceedings of the 58th Annual Meeting of the Association for Computational Linguistics",
    month = jul,
    year = "2020",
    address = "Online",
    publisher = "Association for Computational Linguistics",
    url = "https://aclanthology.org/2020.acl-main.400/",
    doi = "10.18653/v1/2020.acl-main.400",
    pages = "4346--4350"
}

@misc{mma,
title={{MMA}: Benchmarking Multi-Modal Large Language Model in Ambiguity Contexts},
author={Ru Wang and Selena Song and Liang Ding and Shixiang Shane Gu and Mingming Gong and Yusuke Iwasawa and Yutaka Matsuo and Jiaxian Guo},
year={2024},
url={https://openreview.net/forum?id=ywKlmMor0f}
}

@inproceedings{ma-etal-2024-3am,
    title = "3{AM}: An Ambiguity-Aware Multi-Modal Machine Translation Dataset",
    author = "Ma, Xinyu  and
      Liu, Xuebo  and
      Wong, Derek F.  and
      Rao, Jun  and
      Li, Bei  and
      Ding, Liang  and
      Chao, Lidia S.  and
      Tao, Dacheng  and
      Zhang, Min",
    editor = "Calzolari, Nicoletta  and
      Kan, Min-Yen  and
      Hoste, Veronique  and
      Lenci, Alessandro  and
      Sakti, Sakriani  and
      Xue, Nianwen",
    booktitle = "Proceedings of the 2024 Joint International Conference on Computational Linguistics, Language Resources and Evaluation (LREC-COLING 2024)",
    month = may,
    year = "2024",
    address = "Torino, Italia",
    publisher = "ELRA and ICCL",
    url = "https://aclanthology.org/2024.lrec-main.1/",
    pages = "1--13"
}

@article{qwen25,
  title={Qwen2.5 technical report},
  author={{Qwen Team}},
  journal={arXiv preprint arXiv:2412.15115},
  url={https://arxiv.org/abs/2412.15115},
  year={2024}
}

@inproceedings{papineni-etal-2002-bleu,
    title = "BLEU: a Method for Automatic Evaluation of Machine Translation",
    author = "Papineni, Kishore  and
      Roukos, Salim  and
      Ward, Todd  and
      Zhu, Wei-Jing",
    editor = "Isabelle, Pierre  and
      Charniak, Eugene  and
      Lin, Dekang",
    booktitle = "Proceedings of the 40th Annual Meeting of the Association for Computational Linguistics",
    month = jul,
    year = "2002",
    address = "Philadelphia, Pennsylvania, USA",
    publisher = "Association for Computational Linguistics",
    url = "https://aclanthology.org/P02-1040/",
    doi = "10.3115/1073083.1073135",
    pages = "311--318"
}

@inproceedings{rei-etal-2020-comet,
    title = "{COMET}: A Neural Framework for {MT} Evaluation",
    author = "Rei, Ricardo  and
      Stewart, Craig  and
      Farinha, Ana C  and
      Lavie, Alon",
    editor = "Webber, Bonnie  and
      Cohn, Trevor  and
      He, Yulan  and
      Liu, Yang",
    booktitle = "Proceedings of the 2020 Conference on Empirical Methods in Natural Language Processing (EMNLP)",
    month = nov,
    year = "2020",
    address = "Online",
    publisher = "Association for Computational Linguistics",
    url = "https://aclanthology.org/2020.emnlp-main.213/",
    doi = "10.18653/v1/2020.emnlp-main.213",
    pages = "2685--2702"
}

@inproceedings{popovic-2015-chrf,
    title = "chr{F}: character n-gram {F}-score for automatic {MT} evaluation",
    author = "Popovi{\'c}, Maja",
    editor = "Bojar, Ond{\v{r}}ej  and
      Chatterjee, Rajan  and
      Federmann, Christian  and
      Haddow, Barry  and
      Hokamp, Chris  and
      Huck, Matthias  and
      Logacheva, Varvara  and
      Pecina, Pavel",
    booktitle = "Proceedings of the Tenth Workshop on Statistical Machine Translation",
    month = sep,
    year = "2015",
    address = "Lisbon, Portugal",
    publisher = "Association for Computational Linguistics",
    url = "https://aclanthology.org/W15-3049/",
    doi = "10.18653/v1/W15-3049",
    pages = "392--395"
}

@inproceedings{snover-etal-2006-study,
    title = "A Study of Translation Edit Rate with Targeted Human Annotation",
    author = "Snover, Matthew  and
      Dorr, Bonnie  and
      Schwartz, Rich  and
      Micciulla, Linnea  and
      Makhoul, John",
    booktitle = "Proceedings of the 7th Conference of the Association for Machine Translation in the Americas: Technical Papers",
    month = aug # " 8-12",
    year = "2006",
    address = "Cambridge, Massachusetts, USA",
    publisher = "Association for Machine Translation in the Americas",
    url = "https://aclanthology.org/2006.amta-papers.25/",
    pages = "223--231"
}

@inproceedings{devlin2019bert,
  title={{BERT}: Pre-training of deep bidirectional transformers for language understanding},
  author={Devlin, Jacob and Chang, Ming-Wei and Lee, Kenton and Toutanova, Kristina},
  booktitle={Proceedings of the 2019 conference of the North American chapter of the association for computational linguistics: human language technologies, volume 1 (long and short papers)},
  pages={4171--4186},
  year={2019},
  url={https://arxiv.org/abs/1810.04805}
}

@inproceedings{banerjee-lavie-2005-meteor,
    title = "{METEOR}: An Automatic Metric for {MT} Evaluation with Improved Correlation with Human Judgments",
    author = "Banerjee, Satanjeev  and
      Lavie, Alon",
    editor = "Goldstein, Jade  and
      Lavie, Alon  and
      Lin, Chin-Yew  and
      Voss, Clare",
    booktitle = "Proceedings of the {ACL} Workshop on Intrinsic and Extrinsic Evaluation Measures for Machine Translation and/or Summarization",
    month = jun,
    year = "2005",
    address = "Ann Arbor, Michigan",
    publisher = "Association for Computational Linguistics",
    url = "https://aclanthology.org/W05-0909/",
    pages = "65--72"
}

@article{muennighoff2025s1,
  title={s1: Simple test-time scaling},
  author={Muennighoff, Niklas and Yang, Zitong and Shi, Weijia and Li, Xiang Lisa and Fei-Fei, Li and Hajishirzi, Hannaneh and Zettlemoyer, Luke and Liang, Percy and Cand{\`e}s, Emmanuel and Hashimoto, Tatsunori},
  journal={arXiv preprint arXiv:2501.19393},
  url={https://arxiv.org/abs/2501.19393},
  year={2025}
}

@article{bai2025qwen2,
  title={{Qwen2.5-VL} technical report},
  author={Bai, Shuai and Chen, Keqin and Liu, Xuejing and Wang, Jialin and Ge, Wenbin and Song, Sibo and Dang, Kai and Wang, Peng and Wang, Shijie and Tang, Jun and others},
  journal={arXiv preprint arXiv:2502.13923},
  url={https://arxiv.org/abs/2502.13923},
  year={2025}
}

@article{zhu2025internvl3,
  title={{InternVL3}: Exploring Advanced Training and Test-Time Recipes for Open-Source Multimodal Models},
  author={Zhu, Jinguo and Wang, Weiyun and Chen, Zhe and Liu, Zhaoyang and Ye, Shenglong and Gu, Lixin and Tian, Hao and Duan, Yuchen and Su, Weijie and Shao, Jie and others},
  journal={arXiv preprint arXiv:2504.10479},
  url={https://arxiv.org/abs/2504.10479},
  year={2025}
}

@inproceedings{liu-etal-2025-detect,
    title = "Detect, Disambiguate, and Translate: On-Demand Visual Reasoning for Multimodal Machine Translation with Large Vision-Language Models",
    author = "Liu, Danyang  and
      Kong, Fanjie  and
      Sun, Xiaohang  and
      Patil, Dhruva  and
      Vajpayee, Avijit  and
      Liu, Zhu  and
      Bhat, Vimal  and
      Sadoughi, Najmeh",
    editor = "Chiruzzo, Luis  and
      Ritter, Alan  and
      Wang, Lu",
    booktitle = "Proceedings of the 2025 Conference of the Nations of the Americas Chapter of the Association for Computational Linguistics: Human Language Technologies (Volume 1: Long Papers)",
    month = apr,
    year = "2025",
    address = "Albuquerque, New Mexico",
    publisher = "Association for Computational Linguistics",
    url = "https://aclanthology.org/2025.naacl-long.74/",
    doi = "10.18653/v1/2025.naacl-long.74",
    pages = "1559--1570",
    ISBN = "979-8-89176-189-6"
}

@inproceedings{elliott-2018-adversarial,
    title = "Adversarial Evaluation of Multimodal Machine Translation",
    author = "Elliott, Desmond",
    editor = "Riloff, Ellen  and
      Chiang, David  and
      Hockenmaier, Julia  and
      Tsujii, Jun{'}ichi",
    booktitle = "Proceedings of the 2018 Conference on Empirical Methods in Natural Language Processing",
    month = oct # "-" # nov,
    year = "2018",
    address = "Brussels, Belgium",
    publisher = "Association for Computational Linguistics",
    url = "https://aclanthology.org/D18-1329/",
    doi = "10.18653/v1/D18-1329",
    pages = "2974--2978"
}

@inproceedings{wu-etal-2021-good,
    title = "Good for Misconceived Reasons: An Empirical Revisiting on the Need for Visual Context in Multimodal Machine Translation",
    author = "Wu, Zhiyong  and
      Kong, Lingpeng  and
      Bi, Wei  and
      Li, Xiang  and
      Kao, Ben",
    editor = "Zong, Chengqing  and
      Xia, Fei  and
      Li, Wenjie  and
      Navigli, Roberto",
    booktitle = "Proceedings of the 59th Annual Meeting of the Association for Computational Linguistics and the 11th International Joint Conference on Natural Language Processing (Volume 1: Long Papers)",
    month = aug,
    year = "2021",
    address = "Online",
    publisher = "Association for Computational Linguistics",
    url = "https://aclanthology.org/2021.acl-long.480/",
    doi = "10.18653/v1/2021.acl-long.480",
    pages = "6153--6166"
}

@inproceedings{wang-etal-2024-mitigating,
    title = "Mitigating Hallucinations in Large Vision-Language Models with Instruction Contrastive Decoding",
    author = "Wang, Xintong  and Pan, Jingheng  and Ding, Liang  and Biemann, Chris",
    booktitle = "Findings of the Association for Computational Linguistics ACL 2024",
    year = "2024",
    url = "https://aclanthology.org/2024.findings-acl.937",
    pages = "15840--15853",
}

@article{openai2025gpt4o,
  title={{GPT-4o} system card},
  author={Hurst, Aaron and Lerer, Adam and Goucher, Adam P and Perelman, Adam and Ramesh, Aditya and Clark, Aidan and Ostrow, AJ and Welihinda, Akila and Hayes, Alan and Radford, Alec and others},
  journal={arXiv preprint arXiv:2410.21276},
  year={2024}
}

@article{liu2024deepseek,
  title={{DeepSeek-V3} technical report},
  author={Liu, Aixin and Feng, Bei and Xue, Bing and Wang, Bingxuan and Wu, Bochao and Lu, Chengda and Zhao, Chenggang and Deng, Chengqi and Zhang, Chenyu and Ruan, Chong and others},
  journal={arXiv preprint arXiv:2412.19437},
  year={2024},
url={https://arxiv.org/abs/2412.19437}
}

@article{yang2025qwen3,
  title={Qwen3 technical report},
  author={Yang, An and Li, Anfeng and Yang, Baosong and Zhang, Beichen and Hui, Binyuan and Zheng, Bo and Yu, Bowen and Gao, Chang and Huang, Chengen and Lv, Chenxu and others},
  journal={arXiv preprint arXiv:2505.09388},
  year={2025},
url={https://arxiv.org/abs/2505.09388}
}

@article{xing2024mitigating,
  title={Mitigating object hallucination via concentric causal attention},
  author={Xing, Yun and Li, Yiheng and Laptev, Ivan and Lu, Shijian},
  journal={Advances in neural information processing systems},
  volume={37},
  pages={92012--92035},
  year={2024},
  url={https://arxiv.org/abs/2410.15926}
}

@article{chu2025qwen,
  title={Qwen Look Again: Guiding Vision-Language Reasoning Models to Re-attention Visual Information},
  author={Chu, Xu and Chen, Xinrong and Wang, Guanyu and Tan, Zhijie and Huang, Kui and Lv, Wenyu and Mo, Tong and Li, Weiping},
  journal={arXiv preprint arXiv:2505.23558},
  year={2025},
  url={https://arxiv.org/abs/2505.23558}
}

@inproceedings{popovic-2017-chrf,
    title = "chr{F}++: words helping character n-grams",
    author = "Popovi{\'c}, Maja",
    editor = "Bojar, Ond{\v{r}}ej  and
      Buck, Christian  and
      Chatterjee, Rajen  and
      Federmann, Christian  and
      Graham, Yvette  and
      Haddow, Barry  and
      Huck, Matthias  and
      Yepes, Antonio Jimeno  and
      Koehn, Philipp  and
      Kreutzer, Julia",
    booktitle = "Proceedings of the Second Conference on Machine Translation",
    month = sep,
    year = "2017",
    address = "Copenhagen, Denmark",
    publisher = "Association for Computational Linguistics",
    url = "https://aclanthology.org/W17-4770/",
    doi = "10.18653/v1/W17-4770",
    pages = "612--618"
}

@inproceedings{futeral-etal-2023-tackling,
    title = "Tackling Ambiguity with Images: Improved Multimodal Machine Translation and Contrastive Evaluation",
    author = "Futeral, Matthieu  and
      Schmid, Cordelia  and
      Laptev, Ivan  and
      Sagot, Beno{\^i}t  and
      Bawden, Rachel",
    editor = "Rogers, Anna  and
      Boyd-Graber, Jordan  and
      Okazaki, Naoaki",
    booktitle = "Proceedings of the 61st Annual Meeting of the Association for Computational Linguistics (Volume 1: Long Papers)",
    month = jul,
    year = "2023",
    address = "Toronto, Canada",
    publisher = "Association for Computational Linguistics",
    url = "https://aclanthology.org/2023.acl-long.295/",
    doi = "10.18653/v1/2023.acl-long.295",
    pages = "5394--5413"
}

@inproceedings{li-etal-2021-vision,
    title = "Vision Matters When It Should: Sanity Checking Multimodal Machine Translation Models",
    author = "Li, Jiaoda  and
      Ataman, Duygu  and
      Sennrich, Rico",
    editor = "Moens, Marie-Francine  and
      Huang, Xuanjing  and
      Specia, Lucia  and
      Yih, Scott Wen-tau",
    booktitle = "Proceedings of the 2021 Conference on Empirical Methods in Natural Language Processing",
    month = nov,
    year = "2021",
    address = "Online and Punta Cana, Dominican Republic",
    publisher = "Association for Computational Linguistics",
    url = "https://aclanthology.org/2021.emnlp-main.673/",
    doi = "10.18653/v1/2021.emnlp-main.673",
    pages = "8556--8562"
}

@misc{qwen35,
    title = {{Qwen3.5}: Towards Native Multimodal Agents},
    author = {{Qwen Team}},
    year = {2026},
    month = feb,
    url = {https://qwen.ai/blog?id=qwen3.5}
}

@misc{gemma4,
    title = {{Gemma 4} Model Card},
    author = {{Google DeepMind}},
    year = {2026},
    url = {https://huggingface.co/google/gemma-4-26B-A4B-it}
}

@misc{qwen36plus,
    title = {{Qwen3.6-Plus}: Towards Real World Agents},
    url = {https://qwen.ai/blog?id=qwen3.6},
    author = {{Qwen Team}},
    month = {April},
    year = {2026}
}

@article{grattafiori2024llama,
  title={The {Llama} 3 Herd of Models},
  author={Grattafiori, Aaron and Dubey, Abhimanyu and Jauhri, Abhinav and Pandey, Abhinav and Kadian, Abhishek and Al-Dahle, Ahmad and Letman, Aiesha and Mathur, Akhil and Schelten, Alan and Vaughan, Alex and others},
  journal={arXiv preprint arXiv:2407.21783},
  year={2024},
  url={https://arxiv.org/abs/2407.21783}
}

\appendix
\raggedbottom
\setlength{\textfloatsep}{6pt plus 1pt minus 2pt}
\setlength{\floatsep}{6pt plus 1pt minus 2pt}
\setlength{\intextsep}{6pt plus 1pt minus 2pt}

\section{Positioning Against Prior Resources}
\label{sec:positioning}

\begin{table*}[!t]
\centering
\renewcommand{\arraystretch}{1.18}
\resizebox{\textwidth}{!}{%
\begin{tabular}{@{}llll@{}}
\toprule
\textbf{Resource} & \textbf{Generation format} & \textbf{Ambiguity focus} & \textbf{Ambiguity evaluation} \\
\midrule
MLT \citep{lala2018multimodal} & localized lexical translation & lexical choice & rule-based lexical matching \\
\makecell[l]{AmbigCaps / M$^3$-AmbigCaps\\\citep{li-etal-2021-vision,guo-etal-2022-lvp}}
 & open-ended caption translation & primarily gender/pronoun ambiguity & rule-based matching over lexical variants \\
CoMMuTE \citep{futeral-etal-2023-tackling} & contrastive candidate ranking & lexical/gender ambiguity & contrastive accuracy over predefined candidates \\
3AM \citep{ma-etal-2024-3am} & open-ended MMT generation & mainly word-level ambiguity & general MT and ambiguity-oriented evaluation \\
MMA \citep{mma} & VQA-style disambiguation & sentence-level ambiguity & VQA-style answer accuracy \\
\textbf{VIDA} & open-ended MMT generation & lexical, sentence, pragmatic, collective noun & LLM-judge semantic span accuracy \\
\bottomrule
\end{tabular}%
}
\renewcommand{\arraystretch}{1.0}
\caption{Positioning of VIDA relative to prior multimodal disambiguation resources.}
\label{tab:related-positioning}
\end{table*}

Existing resources address visual disambiguation from different angles, but they differ substantially from VIDA in task format and validation target. MLT focuses on localized lexical translation, where the model resolves an ambiguous source word with sentence and image context rather than generating a full translation. AmbigCaps and M$^3$-AmbigCaps use open-ended caption translation, but their ambiguity coverage is concentrated on gender or pronoun choices derived from back-translation. CoMMuTE provides a controlled contrastive setting by asking models to rank predefined translation candidates, which is useful for isolating visual cues but does not evaluate free-form generation. 3AM targets English--Chinese MMT but primarily focuses on word-level ambiguity and includes many instances resolvable from text alone. MMA broadens the phenomenon to sentence-level ambiguity, but is formulated as a VQA task rather than translation. In contrast, VIDA targets open-ended MMT generation, covers word-, sentence-, pragmatic-, and collective-noun ambiguity, and retains only instances whose intended interpretation is verified to require visual evidence.

The evaluation protocols also differ in what they can verify. Contrastive accuracy measures whether a model ranks a predefined correct candidate above distractors, while rule-based matching over lexical variants is tied to expected surface forms. These designs are less suitable for open-ended translation, where correct disambiguation may be expressed through paraphrases or different lexical choices. General MT metrics such as BLEU and COMET remain useful for overall translation quality, but they do not directly verify whether an annotated ambiguous span is resolved correctly, since surface-overlap metrics may penalize valid paraphrases or lexical variation and sentence-level metrics are too coarse-grained for span-level disambiguation. VIDA therefore uses Disambiguation-Centric Metrics to judge span-level semantic resolution while allowing valid paraphrases. \autoref{tab:related-positioning} summarizes the comparison.

\paragraph{Concrete Examples.}
\autoref{tab:positioning_examples} illustrates these differences with examples from each resource. In 3AM, some captions can be translated correctly without the image, while others are paired with an unrelated image. CoMMuTE asks the model to rank two predefined translations rather than generate one, and MMA evaluates visual question answering rather than translation. In contrast, VIDA manually ensures both visual dependency and image--text consistency for every retained instance, requires the model to generate a complete translation, and evaluates whether the image-supported sense is expressed (see \autoref{fig:case_study} for VIDA examples).

\begin{table*}[!t]
\centering
\small
\begin{tabular}{@{}p{0.09\textwidth}p{0.49\textwidth}p{0.36\textwidth}@{}}
\toprule
\textbf{Resource} & \textbf{Example} & \textbf{Limitation for evaluating visual disambiguation} \\
\midrule
3AM & ``A lot of rolls of toilet paper.'' & Correctly translatable from text alone; no visual evidence is needed. \\
3AM & ``A praying mantis perched on a branch at reserve.'' & Paired with an unrelated image (image--text mismatch). \\
CoMMuTE & ``We'll have to get rid of that mole.'' with two predefined French translations: ``Il va falloir enlever ce grain de beauté'' (beauty mark) and ``Il va falloir se débarrasser de cette taupe'' (animal). & The model ranks the two given translations by probability instead of generating a translation. \\
MMA & Caption ``Crying'' with an image and a multiple-choice question about why the person is crying (gold answer: ``Feeling sad''). & Evaluates visual question answering; the caption is not a complete translation input. \\
\bottomrule
\end{tabular}
\caption{Examples illustrating the limitations of prior resources for evaluating visual disambiguation in open-ended MMT.}
\label{tab:positioning_examples}
\end{table*}

\section{Auxiliary Overall Quality Evaluation}
\label{sec:llm_quality}

To complement standard MT metrics, we evaluate overall translation quality with Qwen3.6-Plus \citep{qwen36plus}, a commercial LVLM, as the judge on a 0--100 scale. \autoref{tab:llm_quality_appendix} shows that SFT and CoT-SFT are close in overall quality, and CoT-SFT is often comparable or higher despite stronger disambiguation performance. This supports our main interpretation that CoT-SFT's gains reflect improved ambiguity resolution rather than a severe loss in general translation quality.

\begin{table}[!htbp]
\centering
\resizebox{\columnwidth}{!}{%
\begin{tabular}{llccc}
\toprule
\textbf{Model} & \textbf{Dataset} & \textbf{Vanilla} & \textbf{SFT} & \textbf{CoT-SFT} \\
\midrule
\multirow{4}{*}{InternVL3-8B}
& All-Test & 83.82 & 84.76 & \textbf{86.35} \\
& VIDA-Base-Test & 86.06 & 87.24 & \textbf{87.25} \\
& VIDA-Sent & 79.31 & 78.86 & \textbf{83.57} \\
& VIDA-CollN & 84.26 & 86.34 & \textbf{87.70} \\
\midrule
\multirow{4}{*}{Qwen2.5-VL-7B}
& All-Test & 83.92 & 85.70 & \textbf{85.87} \\
& VIDA-Base-Test & 86.07 & \textbf{87.33} & \textbf{87.33} \\
& VIDA-Sent & 79.10 & 82.24 & \textbf{82.48} \\
& VIDA-CollN & 84.91 & 86.24 & \textbf{86.69} \\
\bottomrule
\end{tabular}%
}
\caption{Auxiliary overall translation quality judged by Qwen3.6-Plus. Scores are on a 0--100 scale and supplement the main metrics in \autoref{tab:combined_comparison}.}
\label{tab:llm_quality_appendix}
\end{table}

\section{Dataset Statistics}
\label{sec:dataset_stat}

\begin{table}[!b]
\centering
\resizebox{\linewidth}{!}{%
\begin{tabular}{lcc}
\toprule
\textbf{Stage} & \textbf{Remaining} & \textbf{Filtered Out} \\
\midrule
Initial raw pairs & 26,452 & -- \\
After image--text filtering & 14,993 & 11,459 \\
After ambiguity verification & 2,500 & 12,493 \\
Human verification coverage & 2,500 (100\%) & 0 \\
\bottomrule
\end{tabular}
}
\caption{Filtering funnel for constructing VIDA.}
\label{tab:pipeline_funnel}
\end{table}

\begin{table}[!b]
\centering
\small
\begin{tabular}{lc}
\toprule
\textbf{Criterion} & \textbf{Cohen's $\kappa$} \\
\midrule
Disambiguation & 0.8033 \\
Fluency & 0.7938 \\
Preservation & 0.8102 \\
\textbf{Average} & \textbf{0.8024} \\
\bottomrule
\end{tabular}
\caption{Inter-annotator agreement for human verification.}
\label{tab:iaa}
\end{table}

\begin{table*}[!t]
\centering
\resizebox{\textwidth}{!}{%
\begin{tabular}{lcccccccccc}
\toprule
\textbf{Model} & \textbf{BLEU} & \textbf{chrF} & \textbf{chrF++} & \textbf{TER} & \textbf{BERT-F1} & \textbf{METEOR} & \textbf{COMET} & \textbf{Disambi-Term} & \textbf{Disambi-Inst.} \\
\midrule
\multicolumn{10}{@{}c@{}}{\textbf{InternVL3-8B vs. Qwen2.5-7B}} \\ \midrule
LVLM               & 48.04 & 41.95 & 32.98 & 40.29 & 86.63 & 58.47 & 84.49 & 50.86 & 39.81 \\
$\uparrow$ \textit{vision }$\uparrow$ & \textit{6.88} & \textit{6.48} & \textit{2.26} & \textit{-5.83} & \textit{2.03} & \textit{6.63} & \textit{2.51} & \underline{\textbf{8.03}} & \underline{\textbf{8.54}} \\
LLM                & 41.16 & 35.47 & 30.72 & 46.12 & 84.60 & 51.84 & 81.98 & 42.83 & 31.27 \\
\midrule
\multicolumn{10}{@{}c@{}}{\textbf{Qwen2.5-VL-7B vs. Qwen2.5-7B}} \\ \midrule
LVLM               & 47.85 & 41.67 & 34.12 & 42.74 & 86.56 & 58.88 & 84.83 & 50.08 & 39.81 \\
$\uparrow$ \textit{vision }$\uparrow$ & \textit{4.27} & \textit{3.59} & \textit{0.89} & \textit{-1.76} & \textit{1.76} & \textit{5.97} & \textit{1.73} & \underline{\textbf{7.25}} & \underline{\textbf{12.18}} \\
LLM                & 41.16 & 35.47 & 30.72 & 46.12 & 84.60 & 51.84 & 81.98 & 42.83 & 31.27 \\
\bottomrule
\end{tabular}%
}
\caption{Ablation results without images on \textbf{All-Test}. The middle row reports the gain from adding the correct visual input over the text-only baseline.}
\label{tab:text_only_ablation}
\end{table*}

\begin{table*}[!t]
\centering
\resizebox{\textwidth}{!}{%
\begin{tabular}{llccccccccc}
\toprule
\textbf{Model} & \textbf{Image Setting} & \textbf{BLEU} & \textbf{chrF} & \textbf{chrF++} & \textbf{TER} & \textbf{BERT-F1} & \textbf{METEOR} & \textbf{COMET} & \textbf{Disambi-Term} & \textbf{Disambi-Inst.} \\
\midrule
\multirow{2}{*}{InternVL3-8B}
& Correct image & 48.04 & 41.95 & 32.98 & 40.29 & 86.63 & 58.47 & 84.49 & 50.86 & 39.81 \\
& Random image  & 44.51 & 38.79 & 32.12 & 41.89 & 85.53 & 54.67 & 82.49 & 42.39 & 30.51 \\
\midrule
\multirow{2}{*}{Qwen2.5-VL-7B}
& Correct image & 47.85 & 41.67 & 34.12 & 42.74 & 86.56 & 58.88 & 84.83 & 50.08 & 39.81 \\
& Random image  & 43.18 & 37.71 & 32.30 & 43.82 & 85.18 & 53.83 & 82.26 & 42.81 & 32.08 \\
\bottomrule
\end{tabular}%
}
\caption{Ablation results with random images on \textbf{All-Test}. Replacing the paired image with a random one substantially degrades both standard MT metrics and disambiguation-centric metrics.}
\label{tab:random_image_ablation}
\end{table*}

\subsection{Definition and Detection of Ambiguous Terms}
\label{sec:ambiguous_term_def}

We define an \textit{ambiguous term} as a word or phrase whose translation admits multiple valid candidates when given only the pure text context. Crucially, its true meaning can only be uniquely determined by referencing the paired visual evidence. Such terms are detected during Stage 1 of our curation pipeline (\autoref{sec:pipeline}): Qwen-Max and DeepSeek-v3 independently evaluate the text-only source for translation uncertainty, and an instance is retained only when both models reach a consensus that ambiguity exists. The two models then extract the specific ambiguous span and produce an \textit{Ambiguity Rationale} that describes the alternative interpretations and identifies the visual cue required for disambiguation. Concrete examples are illustrated in \autoref{fig:case_study}.

\subsection{Pipeline Filtering Statistics}

The large reduction from the raw candidate pool to the final benchmark reflects the strict quality control in our curation pipeline. We begin with 26,452 raw image--text pairs, retain 14,993 after image--text matching and text normalization, and then keep only 2,500 instances after ambiguity detection and dual-LLM verification. The two models show strong bidirectional agreement on the ambiguity decision: 94.38\% of ambiguous candidates flagged by Qwen-Max are independently confirmed by DeepSeek-v3, and 87.50\% in the reverse direction, indicating that the detected ambiguities are robust rather than model-specific artifacts. Subsequent stages refine translations and annotations without discarding additional instances. All 2,500 retained instances are then human-verified by two native Chinese annotators, with substantial inter-annotator agreement across the three review criteria (average Cohen's $\kappa=0.8024$; \autoref{tab:iaa}). The filtering funnel is summarized in \autoref{tab:pipeline_funnel}.

\subsection{Human Verification Protocol}
\label{sec:human_verification}

LLMs are used to screen the source pairs and to draft disambiguated translations, because processing all candidates manually from scratch would be prohibitively costly. The dual-LLM agreement in Stage~1 reduces reliance on a single model, but it does not determine the final ground truth. In Stage~3, the two annotators do not merely accept or reject LLM outputs; they actively rewrite translations that fail any of the three criteria. When one annotator corrects a translation, the other annotator re-verifies the correction. Disagreements are first discussed between the two annotators, and cases that remain unresolved are referred to an additional native Chinese speaker for adjudication.

\subsection{LLM Configurations}

We use GPT-4o (\texttt{gpt-4o-2024-11-20}) with temperature 1.0 and top\_p 1.0. Qwen-Max (\texttt{qwen-max-2025-01-25}) and DeepSeek-V3 are accessed through DashScope with the default decoding parameters (temperature/top\_p of 0.7/0.8 for Qwen-Max and 0.7/0.6 for DeepSeek-V3); no dated snapshot is available for DeepSeek-V3. The maximum number of output tokens is not set for any of these models.

\subsection{VIDA Subset Statistics}

The rigorous pipeline outlined in \autoref{sec:pipeline} results in the construction of \textbf{VIDA} (\textbf{Vi}sually-\textbf{D}ependent \textbf{A}mbiguity), comprising 2,500 visually dependent instances across three subsets: \textbf{VIDA-Base} contains 1,932 samples curated from 3AM and focuses primarily on word-level ambiguities; \textbf{VIDA-CollN} contains 256 collective-noun cases, also curated from 3AM, where visual context concretizes an otherwise abstract group; and \textbf{VIDA-Sent} contains 312 samples adapted from MMA with sentence-level semantic ambiguities. A complete statistical summary is provided in \autoref{tab:vida-subsets}.

\begin{table}[!htbp]
\centering
\resizebox{\linewidth}{!}{%
\begin{tabular}{lcccc}
\toprule
\textbf{Subset} & \textbf{Ambiguity} & \textbf{Size} & \textbf{Avg. Len.} & \textbf{Avg. Ambi.} \\
\midrule
VIDA-Base & Word-level & 1,932 & 11.12 & 1.78 \\
VIDA-Sent & Sentence-level & 312 & 6.00 & 1.00 \\
VIDA-CollN & Word-level & 256 & 10.08 & 1.20 \\
\bottomrule
\end{tabular}
}
\caption{Statistical summary of VIDA subsets.}
\label{tab:vida-subsets}
\end{table}

\autoref{tab:ambi-types} further reports the distribution of ambiguity types in each subset. The types follow the four categories used in Stage~1 ambiguity detection (Prompt~1 in Appendix~\ref{sec:prompts}) and describe the source of ambiguity: \textit{lexical} (multiple senses of a word or phrase), \textit{syntactic} (multiple possible structures), \textit{pragmatic} (context-dependent interpretations), and \textit{cultural/background} (reliance on external cultural or background knowledge). Since an instance can involve more than one type, the labels are multi-label and their counts may exceed the number of instances.

\begin{table}[!htbp]
\centering
\resizebox{\linewidth}{!}{%
\begin{tabular}{lccccc}
\toprule
\textbf{Subset} & \textbf{Size} & \textbf{Lexical} & \textbf{Syntactic} & \textbf{Pragmatic} & \textbf{Cultural} \\
\midrule
VIDA-Base  & 1,932 & 1,309 & 755 & 858 & 163 \\
VIDA-Sent  & 312   & 77    & 64  & 145 & 26 \\
VIDA-CollN & 256   & 253   & 5   & 14  & 2 \\
\bottomrule
\end{tabular}
}
\caption{Distribution of ambiguity types in VIDA subsets (multi-label).}
\label{tab:ambi-types}
\end{table}
\FloatBarrier

\subsection{Visual Dependence}
\label{sec:visual_dependence}

To further verify that VIDA instances require the correct visual context, we evaluate models under two ablated settings on \textbf{All-Test}: (i) \textbf{Text-only}, where the paired image is removed and replaced by the corresponding text-only backbone, and (ii) \textbf{Random Image}, where the correct image is replaced with a randomly sampled image from the test set. Results are shown in \autoref{tab:text_only_ablation} and \autoref{tab:random_image_ablation}. In both cases, performance drops consistently on both standard MT metrics and the proposed disambiguation-centric metrics. These degradations clearly demonstrate the visual dependence of VIDA: the ambiguous text in our dataset must rely on the correct image information to be successfully disambiguated, and performance drops markedly when the model is given no image or a random image.

We further repeat both ablations with the stronger LVLMs from Section~\ref{sec:stronger_models}. For these models, the text-only setting removes the image from the same model. As shown in \autoref{tab:frontier_ablation}, removing or replacing the image reduces All-Test Disambi-Inst.\ by 8.3--10.4 points for both models. The drop is particularly large on VIDA-Sent, e.g., from 57.05 to 35.58 Disambi-Inst.\ when the image is removed for Qwen3.5-27B. This confirms that the visual dependence of VIDA is not specific to smaller models.

\begin{table}[!htbp]
\centering
\resizebox{\columnwidth}{!}{%
\begin{tabular}{llccc}
\toprule
\textbf{Model} & \textbf{Image} & \textbf{COMET} & \textbf{Disambi-Term} & \textbf{Disambi-Inst.} \\
\midrule
\multirow{3}{*}{Qwen3.5-27B} & Correct & 86.38 & 57.25 & 46.47 \\
 & None   & 84.61 & 49.60 & 38.10 \\
 & Random & 84.61 & 47.95 & 36.09 \\
\midrule
\multirow{3}{*}{Gemma-4-26B-A4B} & Correct & 85.53 & 54.86 & 44.46 \\
 & None   & 83.96 & 48.38 & 36.18 \\
 & Random & 83.57 & 46.48 & 34.35 \\
\bottomrule
\end{tabular}%
}
\caption{Visual-dependence ablation with stronger LVLMs on \textbf{All-Test}. \textit{None} removes the image from the same model; \textit{Random} replaces it with a randomly sampled image.}
\label{tab:frontier_ablation}
\end{table}
\FloatBarrier

\section{Judge Reliability and Training Details}
\label{sec:appendix_validation}

\subsection{Judge Reliability}

We further validate the reliability of the Judge from two complementary perspectives. First, we perform \textbf{Cross-Model Judge Verification} by re-evaluating the exact same translation outputs from Qwen2.5-VL-7B (CoT-SFT) on \textbf{All-Test} with a second Judge based on LLaMA3.1-8B \citep{grattafiori2024llama}. As shown in \autoref{tab:cross_model_judge}, the scores remain highly consistent across judge backbones. Second, we measure \textbf{Human-Judge Alignment} on a randomly sampled 20\% subset of VIDA (500 instances) and compute Cohen's Kappa between expert human judgments and the Qwen-based Judge. \autoref{tab:human_judge_alignment} shows strong agreement across all subsets, indicating that the Judge captures semantic disambiguation resolution rather than model-specific preferences.

\begin{table}[!htbp]
\centering
\small
\begin{tabular}{lcc}
\toprule
\textbf{Judge} & \textbf{Disambi-Term} & \textbf{Disambi-Inst.} \\
\midrule
Qwen-Judge   & 55.51 & 46.08 \\
LLaMA-Judge  & 55.72 & 47.69 \\
\bottomrule
\end{tabular}
\caption{Cross-model judge verification on the same Qwen2.5-VL-7B (CoT-SFT) predictions.}
\label{tab:cross_model_judge}
\end{table}

\begin{table}[!htbp]
\centering
\begin{tabular}{lc}
\toprule
\textbf{Dataset} & \textbf{Cohen's Kappa} \\
\midrule
VIDA-Base-Test & 0.8828 \\
VIDA-Sent      & 0.8932 \\
VIDA-CollN     & 0.9023 \\
\textbf{Average} & \textbf{0.8928} \\
\bottomrule
\end{tabular}
\caption{Human-Judge alignment measured by Cohen's Kappa on a 500-instance sample.}
\label{tab:human_judge_alignment}
\end{table}
\FloatBarrier

\subsection{Judge Training Details}

The Judge is fine-tuned from Qwen3-8B as a binary classifier for span-level disambiguation verification. Each training instance contains the source caption, a candidate translation, the annotated ambiguous span, and the gold interpretation of that span; the model predicts whether the ambiguity is correctly resolved in the translation. Gold disambiguated translations are used as positive examples. Negative examples are drawn from candidate translations produced during the curation pipeline that fail to resolve the annotated ambiguity. To reduce false negatives, we further filter out semantically equivalent candidates with an auxiliary LLM-based screening step before training.

We fine-tune the Judge for 10 epochs with learning rate $1\times10^{-5}$ and batch size 8. LoRA is used with rank 8, alpha 32, and dropout 0.1. The judge data contain 1,499 positive and 1,497 negative examples, which are sampled at a positive-to-negative ratio of 2:1 during training. We use a random seed of 42. This contrastive setup encourages the Judge to focus on disambiguation correctness rather than superficial lexical overlap, which is crucial for handling valid paraphrases and lexical variation in open-ended translation. All experiments were conducted on one NVIDIA A100 GPU.

\begin{table*}[!t]
\centering
\resizebox{\textwidth}{!}{%
\begin{tabular}{@{}lllccccccccc@{}}
\toprule
\textbf{Model} & \textbf{Dataset} & \textbf{Setting} & \textbf{BLEU} & \textbf{chrF} & \textbf{chrF++} & \textbf{TER} & \textbf{BERT-F1} & \textbf{METEOR} & \textbf{COMET} & \textbf{Disambi-Term} & \textbf{Disambi-Inst.} \\ \midrule
\multirow{4}{*}{\textbf{Qwen3.5-27B}}
 & All-Test & Vanilla & 48.14 & 41.69 & 33.95 & 40.70 & 87.51 & 58.88 & 86.38 & 57.25 & 46.47 \\
 & VIDA-Base-Test & Vanilla & 50.84 & 44.85 & 36.89 & 37.60 & 89.01 & 63.32 & 87.32 & 68.53 & 55.79 \\
 & VIDA-Sent & Vanilla & 48.66 & 43.14 & 38.95 & 39.53 & 86.66 & 57.39 & 87.32 & 57.05 & 57.05 \\
 & VIDA-CollN & Vanilla & 37.95 & 32.69 & 24.57 & 49.39 & 85.16 & 50.67 & 83.11 & 19.93 & 12.50 \\ \midrule
\multirow{4}{*}{\textbf{Gemma-4-26B-A4B}}
 & All-Test & Vanilla & 45.82 & 39.74 & 34.07 & 42.39 & 86.86 & 57.31 & 85.53 & 54.86 & 44.46 \\
 & VIDA-Base-Test & Vanilla & 48.76 & 43.20 & 37.88 & 38.97 & 88.44 & 61.98 & 86.63 & 65.19 & 52.33 \\
 & VIDA-Sent & Vanilla & 44.37 & 38.94 & 34.88 & 44.38 & 85.49 & 55.28 & 85.72 & 53.21 & 53.21 \\
 & VIDA-CollN & Vanilla & 36.49 & 31.32 & 23.49 & 49.61 & 84.95 & 49.23 & 82.84 & 22.22 & 16.02 \\ \midrule \midrule
\multirow{12}{*}{\textbf{InternVL3-14B}}
 & \multirow{3}{*}{All-Test} & Vanilla & 49.93 & 43.84 & 37.50 & 38.12 & 87.18 & 59.56 & 84.90 & 50.28 & 39.67 \\
 & & SFT & \closehigh{51.23} & \closehigh{45.08} & \closehigh{38.60} & \closehigh{36.95} & 87.67 & 60.48 & 85.75 & 52.97 & 42.63 \\
 & & CoT-SFT & 49.55 & 43.19 & 34.95 & 39.34 & \closehigh{87.69} & \closehigh{60.76} & \closehigh{86.20} & \ambihigh{58.47} & \ambihigh{50.04} \\ \cmidrule(l){2-12}
 & \multirow{3}{*}{VIDA-Base-Test} & Vanilla & 55.53 & 48.83 & 42.55 & 33.28 & 89.38 & 66.06 & 86.35 & 60.71 & 47.59 \\
 & & SFT & \closehigh{57.25} & \closehigh{50.51} & \closehigh{44.06} & \closehigh{31.72} & \closehigh{90.00} & \closehigh{67.39} & \closehigh{87.27} & 63.36 & 51.03 \\
 & & CoT-SFT & 53.31 & 46.47 & 37.89 & 36.04 & 88.94 & 64.98 & 86.42 & \ambihigh{63.75} & \ambihigh{51.90} \\ \cmidrule(l){2-12}
 & \multirow{3}{*}{VIDA-Sent} & Vanilla & 44.93 & 39.57 & 35.43 & 42.25 & 84.72 & 53.88 & 84.79 & 47.44 & 47.44 \\
 & & SFT & 45.08 & 39.49 & 35.55 & \closehigh{41.84} & 84.99 & 54.22 & 85.50 & 51.28 & 51.28 \\
 & & CoT-SFT & \closehigh{46.39} & \closehigh{40.51} & \closehigh{36.28} & 42.28 & \closehigh{86.46} & \closehigh{57.95} & \closehigh{87.12} & \ambihigh{60.58} & \ambihigh{60.58} \\ \cmidrule(l){2-12}
 & \multirow{3}{*}{VIDA-CollN} & Vanilla & 38.78 & 33.55 & 25.16 & 47.44 & 85.18 & 51.73 & 81.74 & 18.36 & 12.16 \\
 & & SFT & 39.76 & 34.44 & 25.83 & 46.68 & 85.62 & 52.42 & 82.61 & 20.00 & 12.94 \\
 & & CoT-SFT & \closehigh{41.99} & \closehigh{36.36} & \closehigh{27.59} & \closehigh{45.62} & \closehigh{86.36} & \closehigh{54.58} & \closehigh{84.59} & \ambihigh{38.69} & \ambihigh{32.94} \\ \bottomrule
\end{tabular}%
}
\caption{Full results of stronger recent LVLMs under vanilla inference (top) and of \textbf{InternVL3-14B} under Vanilla, SFT, and CoT-SFT settings (bottom). In the bottom block, \closehigh{Box} highlights the best standard metrics and \ambihigh{Box} highlights the best disambiguation metrics within each dataset.}
\label{tab:full_stronger_models}
\end{table*}

\section{Statistical Testing Details}
\label{sec:stat_details}

The OOD subsets are relatively small (312 instances in VIDA-Sent and 256 in VIDA-CollN), so differences in disambiguation accuracy between training settings may partly reflect sampling variation. We therefore test whether the improvements of CoT-SFT over SFT on these subsets are statistically reliable. Since both settings are evaluated on the same instances, we use paired analyses.

\paragraph{Disambi-Inst.}
Under each setting, each instance yields a binary correct/incorrect outcome, so the two settings form paired binary observations. We estimate the 95\% confidence interval of the accuracy difference with a paired bootstrap over instances (10,000 resamples), which resamples each instance together with both of its outcomes and makes no distributional assumption about the difference. To test significance, we use the McNemar test, which is designed for paired binary outcomes: instances on which both settings are correct or both are incorrect carry no information about their difference, so the test relies only on the discordant instances where exactly one setting is correct. We use its exact (binomial) version because the number of discordant instances on the small OOD subsets is limited, in which case the chi-squared approximation can be inaccurate.

\paragraph{Disambi-Term.}
The annotated terms within the same instance share the same source sentence, image, and model output, and are therefore not independent. To account for this dependence, we use an instance-level cluster bootstrap that resamples instances together with all of their annotated terms.

\paragraph{Multiple Comparisons.}
Because we perform four comparisons (two models $\times$ two OOD subsets), we additionally apply a Bonferroni correction, giving an adjusted significance threshold of $\alpha=0.05/4=0.0125$.

\paragraph{Notes on \autoref{tab:ood_significance}.}
All differences $\Delta$ in \autoref{tab:ood_significance} are computed from unrounded scores and may therefore differ slightly from the differences of the rounded values in \autoref{tab:combined_comparison}. Since each VIDA-Sent instance contains exactly one annotated span, its term- and instance-level results coincide.

\section{Full Results of Stronger LVLMs and InternVL3-14B}
\label{sec:full_results}

\autoref{tab:full_stronger_models} reports subset-level results for the models in Section~\ref{sec:stronger_models}.
At the subset level, the stronger LVLMs improve over the vanilla 7B/8B models mainly on VIDA-Base-Test and VIDA-Sent, whereas VIDA-CollN remains difficult (12.50 and 16.02 Disambi-Inst.). For InternVL3-14B, CoT-SFT is comparable to SFT on the in-domain VIDA-Base-Test (51.90 vs.\ 51.03 Disambi-Inst.), while yielding large gains on VIDA-Sent (60.58 vs.\ 51.28) and VIDA-CollN (32.94 vs.\ 12.94).

\begin{figure*}[t]
  \centering
  \includegraphics[width=.9\linewidth]{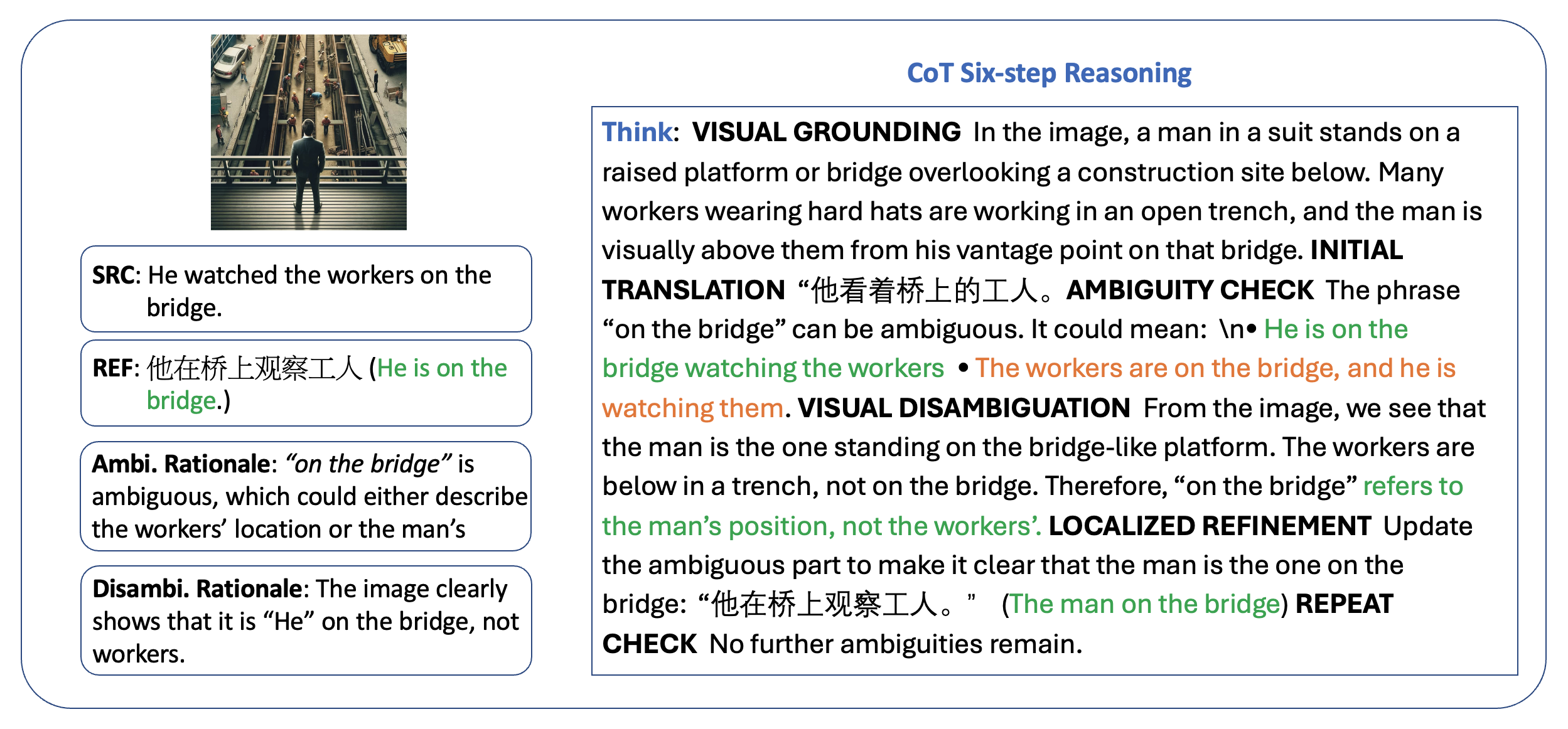}
  \caption{Example of CoT six-step reasoning resolving the ambiguity.}\label{fig:six_step_example}
\end{figure*}
\section{Chain-of-Thought Supervised Fine-Tuning}
\label{sec:cot-sft}

We define a task-specific six-step reasoning template that guides models to articulate the alignment between ambiguous expressions and visual evidence. Each synthetic trace is constructed according to the following standardized reasoning template:
\begin{enumerate}
    \item \textbf{Visual Grounding}: Examine the image carefully and identify the visual elements that correspond to key words or phrases in the source sentence. Describe how these elements connect to the text.
    \item \textbf{Initial Translation}: Generate a preliminary translation based on both the text and the grounded visual evidence.
    \item \textbf{Ambiguity Check}: Review the initial translation and highlight any terms that remain ambiguous—those whose meanings are unclear or context-dependent when relying on text alone.
    \item \textbf{Visual Disambiguation}: This step is critical. While visual grounding establishes a mapping between the image and the text, the initial translation can still leave some ambiguities unresolved. The model explicitly revisits the image, not only to strengthen the connection between ambiguous terms and their corresponding visual evidence, but also to refresh its access to visual information while mitigating the risk of visual token attention decay during long-sequence generation \citep{xing2024mitigating, chu2025qwen} and hallucination \citep{wang-etal-2024-mitigating}. Through this re-examination, the model is better guided to ground its disambiguation decisions in the most relevant visual cues.
    \item \textbf{Localized Refinement}: Update only the ambiguous parts of the initial translation while keeping the rest unchanged. This constraint prevents unnecessary modifications to the sentence structure and helps maintain overall translation fluency.

    \item \textbf{Repeat Check}: Reassess the updated translation. If ambiguities remain, iterate steps 3–5 until the translation is fully disambiguated.
\end{enumerate}

An example is provided in \autoref{fig:six_step_example}.

\begin{figure*}[!t]
  \centering
  \includegraphics[width=.85\linewidth]{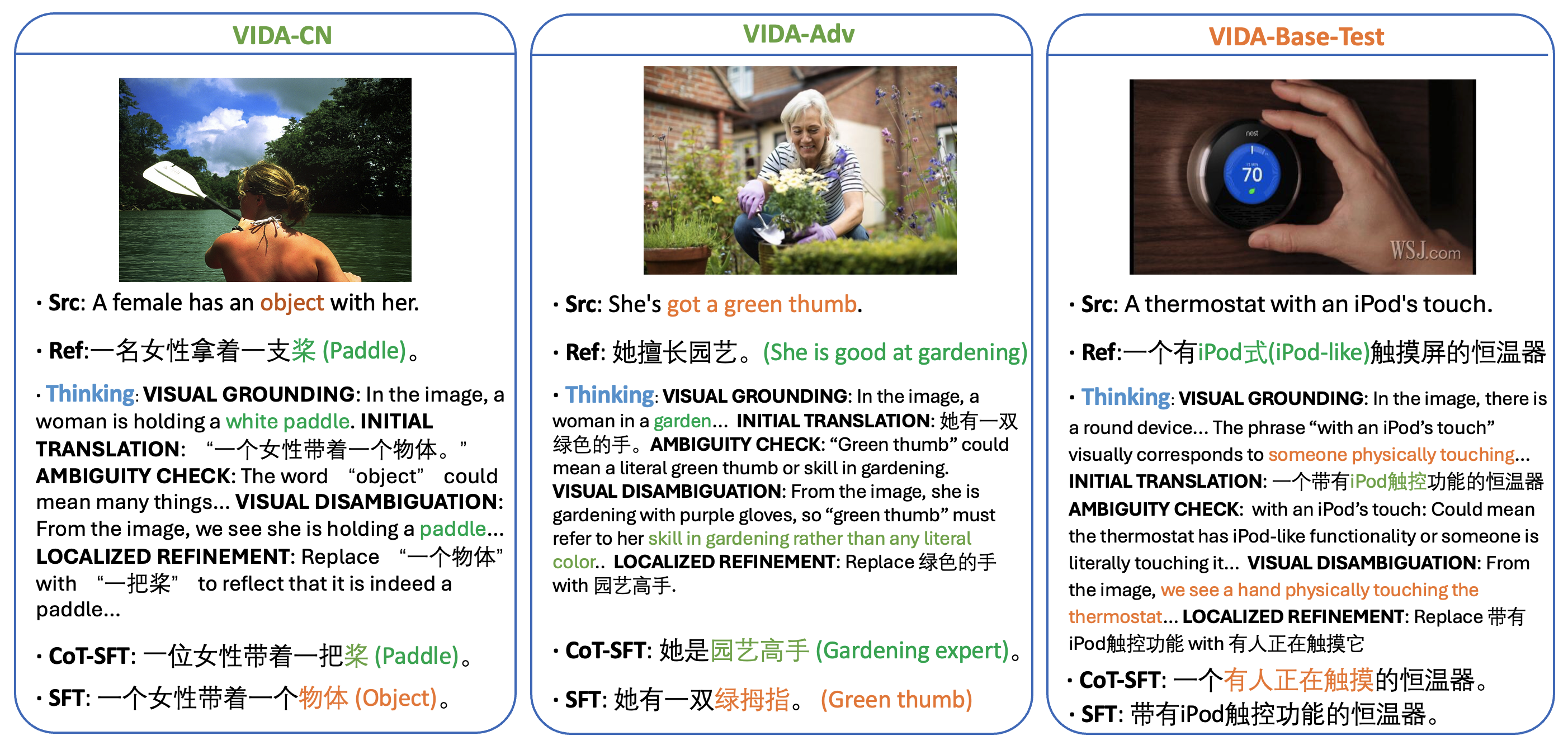}
  \caption{Case study of CoT-SFT vs. SFT}\label{fig:case_study}
\end{figure*}

\section{Error Analysis}
\label{sec:error_analysis}

\subsection{Quantitative Analysis of In-Domain Regressions}
\label{sec:error_analysis_main}

\begin{table}[!htbp]
\centering
\resizebox{\columnwidth}{!}{%
\begin{tabular}{lccccc}
\toprule
\textbf{Model} & \textbf{Cases} & \textbf{Overthinking} & \textbf{Refinement} & \textbf{Grounding} & \textbf{Other} \\
\midrule
Qwen2.5-VL-7B & 62 & 54 (87.1\%) & 5 (8.1\%) & 2 (3.2\%) & 1 (1.6\%) \\
InternVL3-8B & 64 & 55 (85.9\%) & 5 (7.8\%) & 2 (3.1\%) & 2 (3.1\%) \\
\midrule
\textbf{Total} & \textbf{126} & \textbf{109 (86.5\%)} & \textbf{10 (7.9\%)} & \textbf{4 (3.2\%)} & \textbf{3 (2.4\%)} \\
\bottomrule
\end{tabular}%
}
\caption{Error types of in-domain regression cases on VIDA-Base-Test, i.e., instances where SFT resolves all annotated ambiguities but CoT-SFT does not. Percentages are computed within each row.}
\label{tab:error_taxonomy}
\end{table}

Unlike on the OOD subsets, CoT-SFT does not consistently outperform SFT on the in-domain VIDA-Base-Test (\autoref{tab:combined_comparison}): it is better for InternVL3-8B but slightly worse for Qwen2.5-VL-7B. To understand this gap, we analyze where the two settings disagree.

\paragraph{Instance-Level Disagreements.}
Among the 580 test instances, SFT resolves all annotated ambiguities while CoT-SFT does not in 64 instances for InternVL3-8B and 62 for Qwen2.5-VL-7B. Conversely, CoT-SFT is correct while SFT is not in 71 and 48 instances, respectively. The in-domain difference between the two settings is therefore the net result of both gains and regressions, and the regressions are frequent enough to offset most of the gains.

\paragraph{Error Taxonomy.}
To identify the causes of these regressions, we manually analyze all 126 cases where SFT is correct but CoT-SFT is not, inspecting the CoT-SFT reasoning trace together with the image and the source sentence. Following the order of the reasoning template (Appendix~\ref{sec:cot-sft}), we assign each case to one of four mutually exclusive error types:
\begin{itemize}
    \item \textbf{Overthinking}: a correct initial interpretation or translation is later overwritten by an unnecessary or unsupported inference.
    \item \textbf{Refinement error}: the interpretation and reasoning remain correct, but the final localized refinement introduces an omission, unnecessary restructuring, or semantic change.
    \item \textbf{Incorrect visual grounding}: the model misidentifies an object, action, attribute, or relation in the image, leading to an incorrect interpretation or translation.
    \item \textbf{Other translation errors}: residual lexical, grammatical, omission, or realization errors that are not attributable to the preceding categories.
\end{itemize}

\paragraph{Results.}
As shown in \autoref{tab:error_taxonomy}, overthinking accounts for 86.5\% of the regressions (109/126), with similar rates for Qwen2.5-VL-7B (87.1\%) and InternVL3-8B (85.9\%). Refinement errors account for 7.9\%, while incorrect visual grounding (3.2\%) and other translation errors (2.4\%) are rare. The low proportion of grounding errors indicates that the regressions are rarely caused by failing to perceive the image. Instead, the model usually reaches a correct interpretation at an early step but abandons it later in the reasoning process.

\paragraph{Interpretation.}
This pattern explains why the benefit of CoT-SFT is smaller in-domain than under distribution shift. VIDA-Base-Test shares the word-level ambiguity types of the training data, and SFT already fits the reference translation patterns for many of these instances, so a correct translation can often be produced directly. The fixed six-step template nevertheless requires the model to re-examine every instance, and each additional step provides an opportunity to overwrite an adequate initial translation with spurious evidence, such as irrelevant visual details or an overly literal reading of an idiomatic expression (see the example in Appendix~\ref{sec:error_qualitative}). On the OOD subsets, where direct translation fails more often, the same re-examination is more likely to correct errors than to introduce them. These findings suggest that adapting the reasoning depth to the difficulty of the ambiguity, for example, by stopping after the ambiguity check when no unresolved ambiguity is detected, is a promising direction for reducing overthinking.

\subsection{Qualitative Examples}
\label{sec:error_qualitative}

The OOD results in \autoref{sec:experiments} raise a key question: \textit{how does explicit reliance on visual information shape the model's reasoning?} \autoref{fig:case_study} (left and middle) illustrates how CoT-SFT aligns ambiguous terms with visual evidence in \textbf{VIDA-CollN} and \textbf{VIDA-Sent}.

The \textbf{VIDA-CollN} example (left of \autoref{fig:case_study}) illustrates the collective noun ambiguity: the source sentence contains the ambiguous noun "object", which requires a concrete translation ("paddle"). The SFT model, without reasoning, outputs the literal "object," which fails to capture the intended meaning. In contrast, the CoT-SFT shows that the model first generates an initial translation \mbox{(\begin{CJK}{UTF8}{gbsn}物体\end{CJK})}, maintaining the literal meaning. During the ambiguity check, the model detects that "object" is ambiguous. In the subsequent visual disambiguation step, it grounds the word to the image and identifies that the woman is holding a paddle. Finally, in localized refinement, the model updates the translation to "paddle", producing the correct disambiguated output.

The \textbf{VIDA-Sent} example (middle of \autoref{fig:case_study}) demonstrates sentence-level ambiguity where an idiomatic expression could be misunderstood literally. The phrase "got a green thumb" could be interpreted literally or idiomatically. The SFT model again produces a literal Chinese rendering of the thumb color. In contrast, the CoT-SFT first provides a literal initial translation \mbox{(\begin{CJK}{UTF8}{gbsn}绿色的手\end{CJK})}. Through visual disambiguation, it recognizes from the image that the woman is gardening, and therefore refines the output to "gardening expert", correctly capturing the idiomatic meaning. 


Although CoT-SFT can improve disambiguation accuracy, we observe that it may also introduce \emph{overthinking} on relatively straightforward inputs, which can degrade translation quality. In our reasoning template, this behavior often arises after the model produces an adequate initial translation: subsequent reasoning steps may overwrite the initial output by injecting
unnecessary or spurious reasoning, e.g., overusing irrelevant visual details or interpreting idiomatic expressions too literally, thereby leading to flawed revisions.

The right panel of \autoref{fig:case_study} illustrates an overthinking case. The phrase \emph{"iPod’s touch"} should be interpreted as \emph{"iPod-like touch screen"}. The model first provides a reasonable image description and recognizes the intended interpretation during ambiguity checking. However, in the later disambiguation step, it over-interprets the phrase by incorrectly linking it to \emph{"someone physically touching"} mentioned in the grounding step, rather than the relevant cue about the product feature. As a result, the model revises an initially adequate interpretation into an incorrect final translation. This example suggests that excessive reasoning can override correct early hypotheses and illustrates the dominant in-domain error type quantified in Appendix~\ref{sec:error_analysis_main}.

\onecolumn
\section{Prompts}
\label{sec:prompts}

This section lists the prompts used in VIDA construction and disambiguation evaluation.

\begin{promptbox}[title={Prompt 1: Stage 1 Ambiguity Detection}]
You are a professional text analysis and disambiguation system specialized in English-to-Chinese translation.
Your task is to:
1. Determine whether a given English caption contains any ambiguity when interpreted without any additional context or images.
2. If ambiguity exists, explain:
   - The type of ambiguity (lexical, syntactic, pragmatic, or cultural/background).
   - The reason it is ambiguous (how multiple interpretations can arise).
   - Potential different Chinese translations reflecting these interpretations.
3. If no ambiguity exists, respond with that conclusion.

Ambiguity Definition:
- Lexical: a word or phrase has multiple meanings (e.g., "bank" = financial institution vs. river bank).
- Syntactic: the sentence structure permits multiple interpretations (e.g., "I saw the man with a telescope").
- Pragmatic: the context or speaker's intention is unclear (e.g., "Don't put all your eggs in one basket" may be literal or metaphorical).
- Cultural/Background: specialized or culturally dependent knowledge is required (e.g., "Break a leg!").

Input:
- A single English caption (no images, no external context).

Output Requirements:
Return a JSON object with the following structure:
{
  "has_ambiguity": <true | false>,
  "ambiguities": [
    {
      "type": "<lexical | syntactic | pragmatic | cultural-background>",
      "explanation": "<why it is ambiguous>",
      "possible_chinese_translations": [
        "<translation option 1>",
        "<translation option 2>"
      ],
      "ambiguous_terms": ["<term1>"]
    }
  ]
}
If the caption is unambiguous:
{ "has_ambiguity": false, "ambiguities": [] }

Important Notes:
- Only consider the caption's text itself (no external context or visual cues).
- If multiple ambiguity types exist, list them all.
- The "possible_chinese_translations" field should illustrate how each interpretation would be rendered differently in Chinese.
- Be concise but clear.

Example:
Caption: "Jaguar is approaching rapidly."
{
  "has_ambiguity": true,
  "ambiguities": [
    {
      "type": "lexical",
      "explanation": "The word 'Jaguar' can refer to an animal or a car brand.",
      "possible_chinese_translations": [
        "A jaguar is approaching rapidly.",
        "A Jaguar car is approaching rapidly."
      ],
      "ambiguous_terms": ["Jaguar"]
    }
  ]
}
Now, please analyze the next English caption according to these instructions.
\end{promptbox}

\clearpage
\begin{promptbox}[title={Prompt 2: Stage 1 Consensus Merging}]
You are a professional data processing assistant skilled in understanding English ambiguities and structuring annotation data. Given ambiguity annotations from two independent models (Qwen-Max and DeepSeek-v3) for the same English caption, merge them into a single structured record.

Processing Rules:
1. Merge ambiguities by type:
   - Group ambiguity entries from both qwen_ambi and v3_ambi by their "type" field (lexical, syntactic, pragmatic, cultural/background).
   - If two ambiguities from different models share the same type and describe the same underlying issue, merge them.
   - If ambiguities differ substantially even under the same type, keep them separate.
2. Merging Details:
   - explanation: combine the explanations from both sources into a single concise paragraph.
   - translations: union the translation candidates from both sources, removing exact duplicates.
3. Extract Ambiguous Terms:
   - From the original English sentence (the "en" field), extract the literal word(s) or phrase(s) that cause each ambiguity.
   - Save them into a new field "ambiguous_terms" (a list).
   - Terms must be taken literally from the original English sentence.

Output Format:
[
  {
    "type": "lexical",
    "explanation": "<combined explanation>",
    "translations": [
      "<merged translation 1>",
      "<merged translation 2>"
    ],
    "ambiguous_terms": ["<term1>", "<term2>"]
  },
  {
    "type": "syntactic",
    "explanation": "<combined explanation>",
    "translations": [...],
    "ambiguous_terms": [...]
  }
]
Begin processing based on these instructions.
\end{promptbox}

\begin{promptbox}[title={Prompt 3: Stage 2 Disambiguation Translation}]
You are an English-to-Chinese multimodal translation expert.

Input:
1. A single English caption (text).
2. One image showing the real-world scene the caption describes.
3. A list of ambiguity notes generated by a previous model.

Your task:
- Look at BOTH the text and the image.
- Disambiguate the caption and produce the most accurate, fluent Chinese translation.
- Briefly state which ambiguity was resolved by the visual evidence.

Output JSON (Chinese UTF-8):
{
  "translation_zh": "<final Chinese translation>",
  "resolved_ambiguity": "<brief explanation of which type of ambiguity was resolved and how the image disambiguated it>"
}

Rules:
- If the caption is actually unambiguous even without the image, set "resolved_ambiguity" to "N/A".
- Do not describe the image; focus on producing the best translation.
- Use concise, standard-register Simplified Chinese with natural and fluent word order.

User-message template:
Caption: "<english_caption>"
Ambiguity notes:
<merged ambiguity records from Stage 1>
\end{promptbox}

\begin{promptbox}[title={Prompt 4: Disambiguation Judge}]

You are an English-to-Chinese translation reviewer. Your task is to judge whether a given Chinese translation correctly conveys a specified "gold sense" (the intended meaning of an ambiguous term in context).

Inputs:
1. The English source sentence.
2. The Chinese translation under evaluation.
3. The ambiguous term or phrase in the source (ambi_term).
4. The gold sense (gold_sense), expressed in Chinese, describing the meaning we expect the ambiguous term to convey in the given context.
5. A reference Chinese translation.

Task:
Based on the gold sense (4) and the reference translation (5), judge whether the Chinese translation under evaluation (2) accurately expresses this meaning.
- If yes, return "Correct".
- If the meaning is missing or distorted, return "Incorrect".

Output format (strictly two lines):
Correct/Incorrect, brief reason

English source: {en}
Chinese translation: {output}
Ambiguous term: {ambi_term}
Gold sense: {gold_sense}
Reference translation: {reference_translation}
\end{promptbox}

\end{document}